\pdfoutput=1
\documentclass[runningheads]{llncs}
\usepackage{graphicx}
\usepackage{amsmath,amssymb} 
\usepackage{color}

\usepackage{subfigure}
\usepackage{multirow}
\usepackage{tabu}
\usepackage[width=122mm,left=12mm,paperwidth=146mm,height=193mm,top=12mm,paperheight=217mm]{geometry}
\begin{document}
\pagestyle{headings}
\mainmatter

\title{On Duality Of Multiple Target Tracking and Segmentation} 

\titlerunning{On Duality Of Multiple Target Tracking and Segmentation}

\authorrunning{Yicong Tian, Mubarak Shah}

\author{Yicong Tian\hspace{0.5 in} Mubarak Shah}


\institute{Center for Research in Computer Vision, University of Central Florida\\
	\email{ \{ytian,shah\}@crcv.ucf.edu}
}

\maketitle

\begin{abstract}
	Traditionally, object tracking and segmentation are treated as two separate problems and solved independently. However, in this paper, we argue that tracking and segmentation are actually closely related and solving one should help the other. On one hand, the object track, which is a set of bounding boxes with one bounding box in every frame, would provide strong high-level guidance for the target/background segmentation task. On the other hand, the object segmentation would separate object from other objects and background, which will be useful for determining track locations in every frame. We propose a novel framework which combines online multiple target tracking and segmentation in a video. In our approach, the tracking and segmentation problems are coupled by Lagrange dual decomposition, which leads to more accurate segmentation results and also \emph{helps resolve typical difficulties in multiple target tracking, such as occlusion handling, ID-switch and track drifting}. To track targets, an individual appearance model is learned for each target via structured learning and network flow is employed to generate tracks from densely sampled candidates. For segmentation, multi-label Conditional Random Field (CRF) is applied to a superpixel based spatio-temporal graph in a segment of video to assign background or target labels to every superpixel. The experiments on diverse sequences show that our method outperforms state-of-the-art approaches for multiple target tracking as well as segmentation.
\end{abstract}

\section{Introduction}
Segmentation is a classic problem in Computer Vision through which foreground and background can be separated in an {\em image}. Numerous edge detection and region segmentation based methods have been proposed for more than fifty years. Compared to a single image, video provides additional temporal information for solving segmentation problems in multiple frames of video. The goal of video object segmentation is to assign foreground labels to pixels corresponding to each target of interest while assigning background labels to remaining pixels. Video object segmentation has several applications including activity understanding, content based video retrieval, video summarization, etc. Closely related to video segmentation is multiple object tracking, where the aim is to determine the corresponding locations of all targets in every frame of a video, with applications such as video surveillance, human-computer interaction, anomaly detection, etc. In this paper, we argue that tracking and segmentation are actually closely related and solving one should help the other.

The track of a target is typically represented as a sequence of bounding boxes enclosing the target. Though convenient, bounding boxes are coarse approximations of targets. Moreover, since bounding boxes usually include non-target pixels, the features extracted from them could be contaminated by background pixels. When these features are used as target representation in tracking, they may cause drift, ID-switches and inaccurate target localizations. Therefore, the ultimate goal of tracking should be to determine the locations of target with corresponding foreground pixels (regions) instead of just coarse bounding boxes.

Traditionally, object tracking and segmentation have been treated as two separate problems and solved independently. However, we argue that tracking and segmentation are actually closely related and solving them should help each other. On one hand, the object track, which is a set of bounding boxes with one bounding box in every frame, would provide strong high-level guidance for the target/background segmentation task. Pixels within a target box are highly likely to be labelled as the target. Conversely, the chance that pixels far away from the box belonging to the target is quite low. On the other hand, the object segmentation would separate object from other objects and background, which will be useful for determining track locations in every frame. This will help in resolving common issues in tracking. For example, during occlusion, the bounding box based appearance score of the occluded target is typically low, posing difficulty in tracking. However, the pixel labels in the visible part of the target would guide tracker to find the correct location of the target. In addition, pixels' target/background labels contain information about target identities and locations, thus will help in avoiding track drifting and ID-switches.

In this paper, we propose to combine online multiple target tracking and segmentation in one framework. The key idea to couple these two tasks is that \emph{the target/background pixel labels are highly correlated with target boxes}. In other words, target pixels are more likely to fall in the target's corresponding box and object boxes should contain as less background pixels as possible. To solve the combined problem, Lagrange dual decomposition is employed. In each iteration, the two subproblems are solved independently with the Lagrange variables serving as a connection between them. Instead of pre-trained object detector, online learned discriminative appearance models are used for the tracking subproblem. The appearance models are learned by structured learning which adapt themselves during tracking. Tracks are then generated by solving multiple network flow problems, one for each target. For the segmentation subproblem, a foreground Gaussian Mixture Model (GMM) is constructed for each target along with one universal background GMM. These GMMs are used to compute target/background confidence maps. For a segment of video (a few frames), a superpixel based spatio-temporal graph is built and multi-label CRF is applied to the graph to obtain final target/background labeling.

\begin{figure}[t]
	\centering
	\subfigure[]{\label{fig:fig3:a}
		\includegraphics[width=\textwidth]{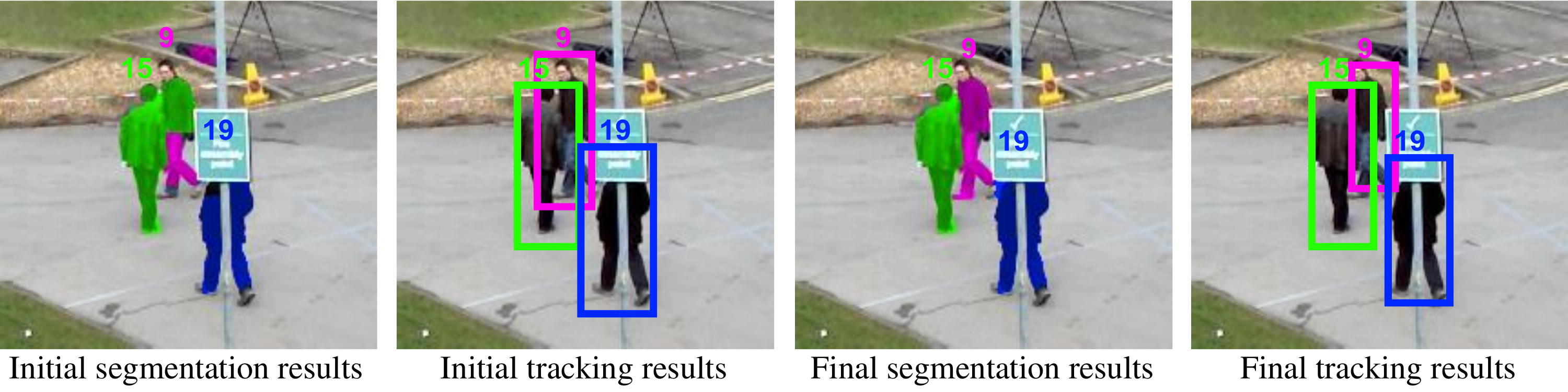}}
	\subfigure[]{\label{fig:fig3:b}
		\includegraphics[width=\textwidth]{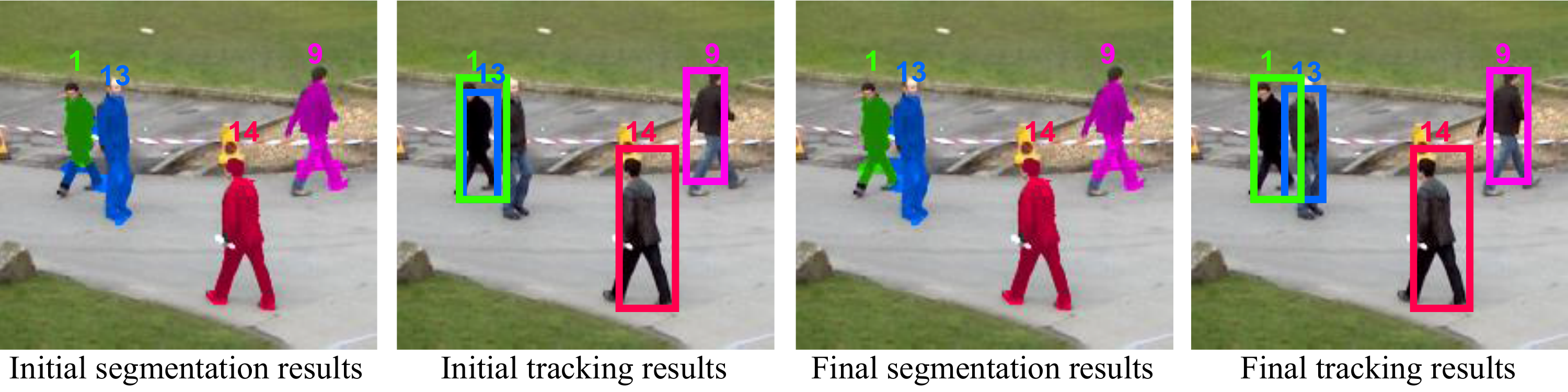}}
	\caption{Two examples of the tracking and segmentation tasks benefiting from each other (zoomed in views are shown). (a) By applying pure segmentation, the upper body of target $9$ is mislabelled as target $15$ due to similar color. But the tracking part is able to track target $9$ correctly. After dual decomposition, the whole body of target $9$ is labelled correctly and more accurate box is obtained for target $9$. (b) Without incorporating segmentation, the track for target $13$ drifts to target $1$. However, the segmentation results for target $13$ are correct using pure segmentation. After dual decomposition, target $13$ is tracked successfully and the segmentation results for target $1$ are also improved. Combining the two subproblems lead to both better tracking and better segmentation results.}
	\label{fig:fig3}
\end{figure}

This paper makes three important contributions. First, we propose a novel framework which combines multiple target tracking and segmentation in one energy function. The two tasks benefit from each other, thus leading to both better tracking and better segmentation results (See Fig. \ref{fig:fig3}). Second, the unified energy function is optimized effectively using dual decomposition. Third, the proposed approach is able to track multiple targets in terms of finer segments (regions) supported by corresponding target pixels rather than coarse bounding boxes.

\section{Related Work}

\subsection{Multiple target tracking}
Most approaches \cite{zamir2012gmcp,zhang2008global,pirsiavash2011globally} for multiple target tracking follow tracking-by-detection framework. First, a pre-trained object detector is applied to find a set of candidate locations for targets. Then these candidates are fed into a data association mechanism to form tracks. Though these methods show competitive tracking results, their performance heavily depend on object detector outputs, which are usually poor when dealing with occlusion and articulated objects. In some recent approaches \cite{zhang2013structure,dehghan2015target}, target-specific appearance models are trained where detection and tracking are combined in one framework. The models adapt to target appearance change during tracking. Inspired by these methods, we learn target-specific appearance models instead of using a generic object detector to better handle cases where a generic object detector does not perform well.

The data association step in tracking-by-detection approaches assigns candidate locations to different targets in the scene and it has been formulated as a network flow problem recently \cite{zhang2008global,pirsiavash2011globally,berclaz2011multiple,butt2013multi}. The data association step is important because the candidate locations are generated by a generic object detector which has no notion of target identities. In contrast, the appearance models used in our framework are target-specific and the generated candidate locations already include target identity information, so we construct a network and find the min-cost flow for each target separately to select the best candidate locations.

\subsection{Object Segmentation in Video}
Video object segmentation \cite{brox2010object,ma2012maximum,zhang2013video} aims to segment foreground pixels belonging to the object from the background in every frame. It has been used in combination with single object tracking in \cite{yin2009shape,li2013video,wen2015jots}. However, the videos they use typically contain only one or two main moving objects. Different from these approaches, we solve segmentation along with multiple target tracking. The goal is to segment multiple interacting targets and preserve targets' identities at the same time. Authors in \cite{bibby2010real,horbert2011level,mitzel2010multi} track contours of targets using a level-set framework. Milan \emph{et al.} \cite{milan2015joint} propose a CRF model to jointly optimize over tracking and segmentation. First, a large number of trajectory hypotheses are generated by two trackers (\cite{pirsiavash2011globally} and \cite{henriques2012exploiting}) using human detection results. Then the objective becomes assigning detections and superpixels to trajectory hypotheses. In contrast, we propose an energy function coupling the tracking and segmentation subproblems, which is solved using dual decomposition by taking advantage of synergies between them. In addition, we do not rely on human detection or other trackers.

\subsection{Dual Decomposition}
Dual decomposition is a general and powerful technique widely used in optimization. It solves a problem by decomposing the original problem into multiple subproblems, solving the subproblems separately and then merging the solutions to solve the overall problem. Using dual decomposition, Strandmark and Kahl \cite{strandmark2010parallel} solve the max-flow/min-cut problem in parallel by splitting a large graph into multiple subgraphs. Thus the algorithm runs much faster when multiple CPU cores are available and is also able to handle graph that is too large to fit in computer's RAM. Wu \emph{et al.} \cite{wu2012coupling} propose to incorporate both object detection and data association in a single objective function to avoid error propagation. The objective function is optimized by dual decomposition. Wang and Koller \cite{wang2011multi} construct a unified model over human poses as well as pixel-wise foreground/background segmentation and optimize the energy function using dual decomposition. To the best of our knowledge, we are the first ones to utilize dual decomposition to solve the multiple target tracking and segmentation problems.  

\section{Problem Formulation} \label{sec:subproblems}
In this section, we formulate the two subproblems coupled in our dual decomposition framework: the multiple target tracking problem in Sec. \ref{sec:track} and the target segmentation problem in Sec. \ref{sec:seg}.

\subsection{Multiple Target Tracking} \label{sec:track}
In traditional tracking-by-detection approaches, a set of detections, obtained by a generic object detector, serve as candidate boxes during tracking. Though the performance of object detectors have been improved remarkably in recent years, miss-detections and false alarms still occur quite often in realistic videos, especially in cases of occlusion and articulated objects, thus impacting the tracking performance.  Therefore, we formulate the multiple target tracking in an online manner and learn target appearance models through structural SVM.

To learn an appearance model for target $i$, we assume its bounding box location $\bar{y}_i$ in its first frame $x_i$ is given. We cast this task as a maximum margin structured learning problem~\cite{tsochantaridis2005large} and the objective function is
\begin{equation}
\label{eq:SSVM_ObjectiveFunction}
\centering
\begin{split}
& \hspace{4cm} \min_{\mathbf{w}_i}\frac{1}{2}\lVert \mathbf{w}_i\rVert^2+C\xi, \\
& \hspace{0.5cm} s.t. \quad \xi\geq 0, \quad \mathbf{w}_i\phi (x_i,\bar{y}_i)-\mathbf{w}_i\phi (x_i,y_i)\geq\Delta(\bar{y}_i,y_i)-\xi \quad \forall y_i \in \mathcal Y\setminus \bar{y}_i
\end{split}
\end{equation}
\noindent where $\mathbf{w}_i$ is the learned appearance model for target $i$ and $\phi(x_i,y_i)$ represents the combination of HOG feature \cite{dalal2005histograms} and color histogram \cite{domke2006deformation} extracted at location $y_i$ in frame $x_i$. $\bar{y}_i$ is the ground truth bounding box and $y_i$ is any possible bounding box in the search space other than $\bar{y}_i$. $\Delta(\bar{y}_i,y_i)=(1-(\bar{y}_i\cap y_i))$ defines the loss function if $y_i$ is the predicted bounding box when the ground truth is $\bar{y}_i$ based on the overlap between $\bar{y}_i$ and $y_i$. The popular cutting plane strategy~\cite{tsochantaridis2005large} is employed to solve the structured learning problem. To deal with gradual changes in target appearances, the appearance models are updated using passive-aggressive algorithm \cite{crammer2006online} during tracking.

Based on the tracking results in the previous frame, we densely sample bounding boxes around each target location as candidate locations for frames in current segment of video. Candidate bounding boxes are sampled at three scales, i.e. 0.95, 1 and 1.05 times of the previous box size, since target size may change gradually in the video.

Given all the candidate boxes in current segment of video, for target $i$, the goal of tracking is to select one box in every frame to form a track. We follow the network flow formulation \cite{zhang2008global,pirsiavash2011globally} to solve this problem. A graph $G=(V,E)$ is constructed for target $i$ and the goal of network flow problem is to minimize the following energy function
\begin{equation}
\label{eq:network_flow}
\centering
\begin{split}
E_{track}(Y) =\sum_{n}c_n^sy_n^s+\sum_{mn\in E}c_{mn}y_{mn} +\sum_{n}c_ny_n+\sum_{n}c_n^ty_n^t, \\
s.t. \quad y_{mn},y_n,y_n^s,y_n^t\in\{0,1\}, \quad y_n^s+\sum_{m}y_{mn}=y_n=y_n^t+\sum_{m}y_{nm},
\end{split}
\end{equation}
where $Y$ denotes which candidate boxes are selected. $y_n$ and $y_{mn}$ are binary values, where $y_n=1$ indicates box $n$ is selected and $y_{mn}=1$ means box $m$ and box $n$ from two consecutive frames are both selected. $c_n$ denotes the cost of selecting box $n$ in the track. It is computed by applying the appearance model $\mathbf{w}_i$ to box $n$, which measures how likely the hypothesis box corresponds to target $i$. $c_{mn}$ denotes the cost of including both candidate boxes $m$ and $n$ from two consecutive frames in the track. It is computed as the histogram intersection between the color histograms extracted from the two boxes. $c_n^s$ and $c_n^t$ represent the costs to start and end a track respectively. They are both set to a fixed number. The goal of the network flow problem is to select a track such that the costs of selected boxes are minimized, which is solved efficiently using dynamic programming. By solving a network flow problem with corresponding candidate boxes for each target, the tracks of all targets are obtained.

\subsection{Target Segmentation} \label{sec:seg}
Segmentation aims to find foreground pixels corresponding to each target so that precise object contour can be determined, instead of typical bounding box representation. In this section, we describe the procedure to get foreground/background segmentation for all targets in a segment of video.

\begin{figure}[t]
	\centering
	\subfigure[]{\label{fig:fig2:b}
		\includegraphics[height=1.35cm]{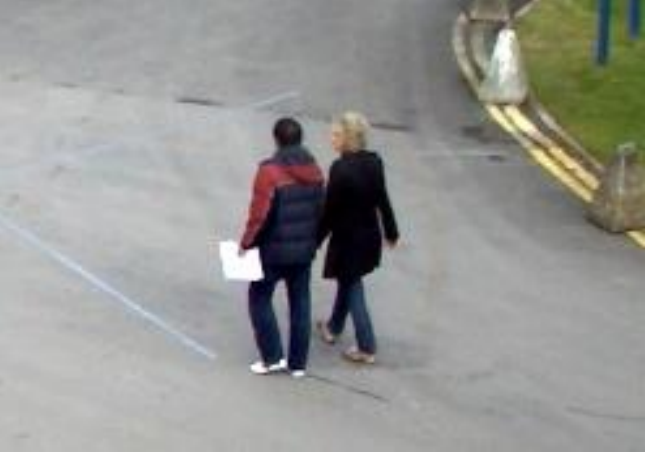}}%
	\subfigure[]{\label{fig:fig2:c}
		\includegraphics[height=1.35cm]{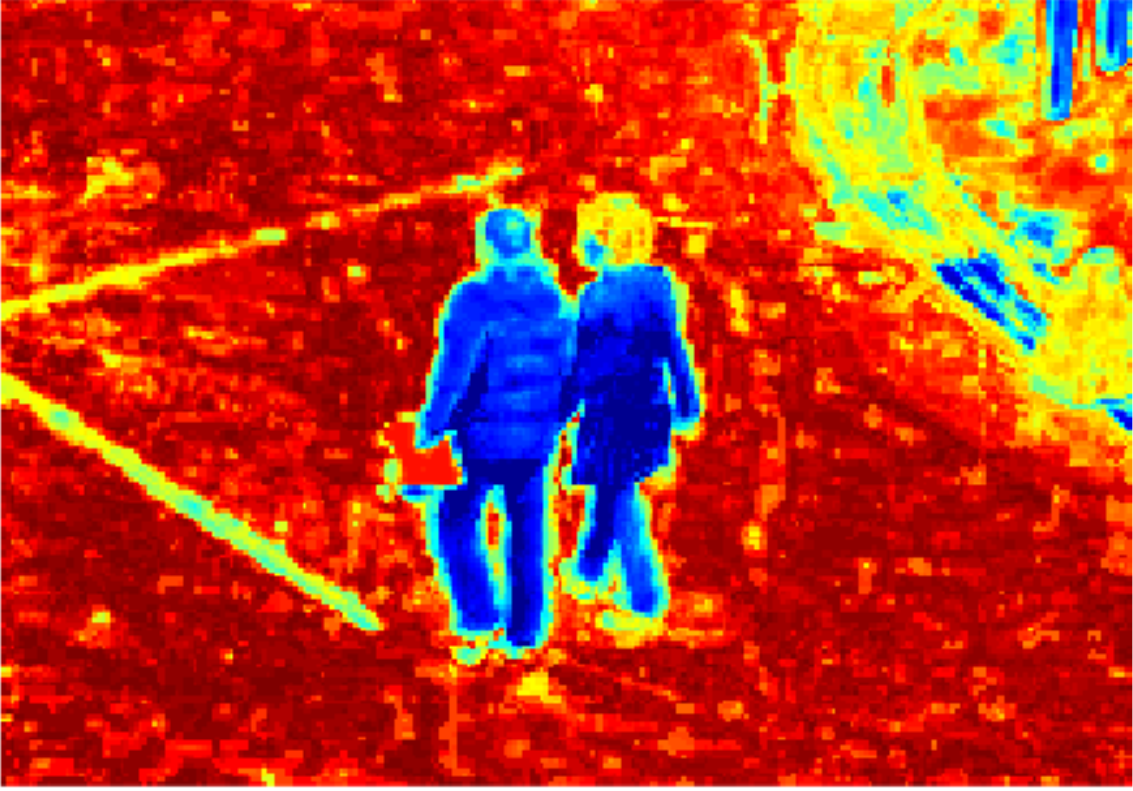}}%
	\subfigure[]{\label{fig:fig2:d}
		\includegraphics[height=1.35cm]{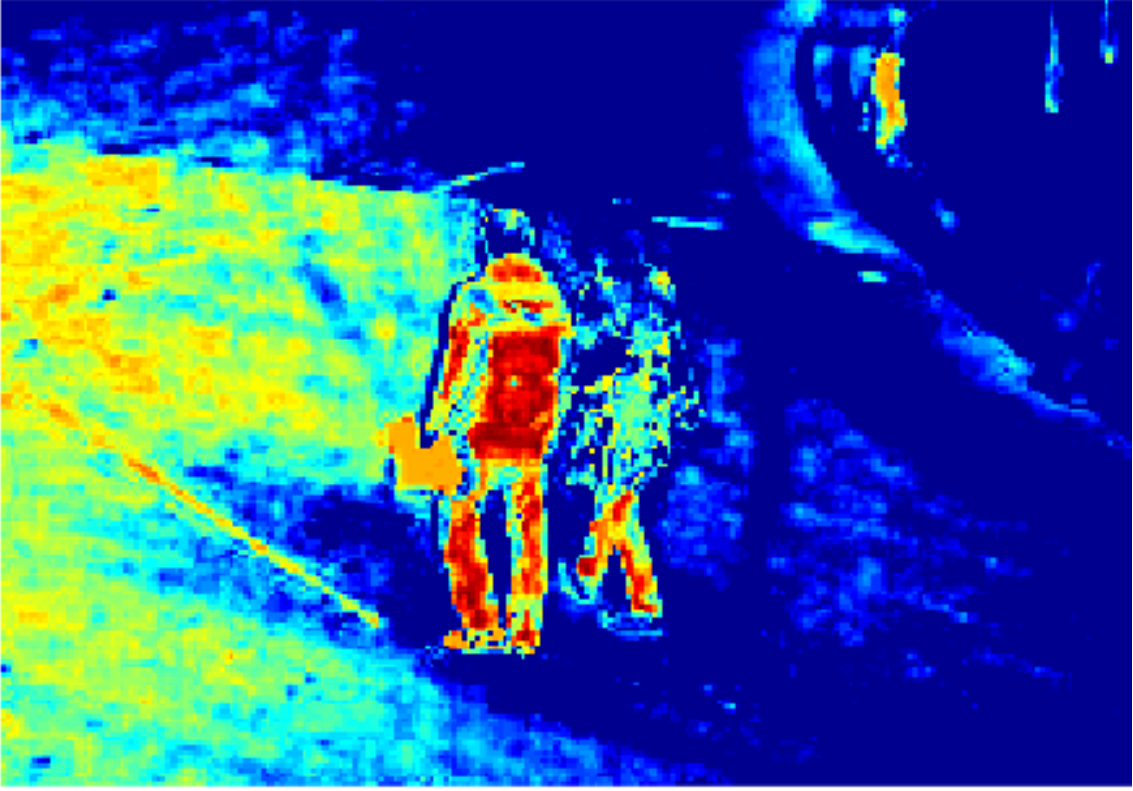}}%
	\subfigure[]{\label{fig:fig2:e}
		\includegraphics[height=1.35cm]{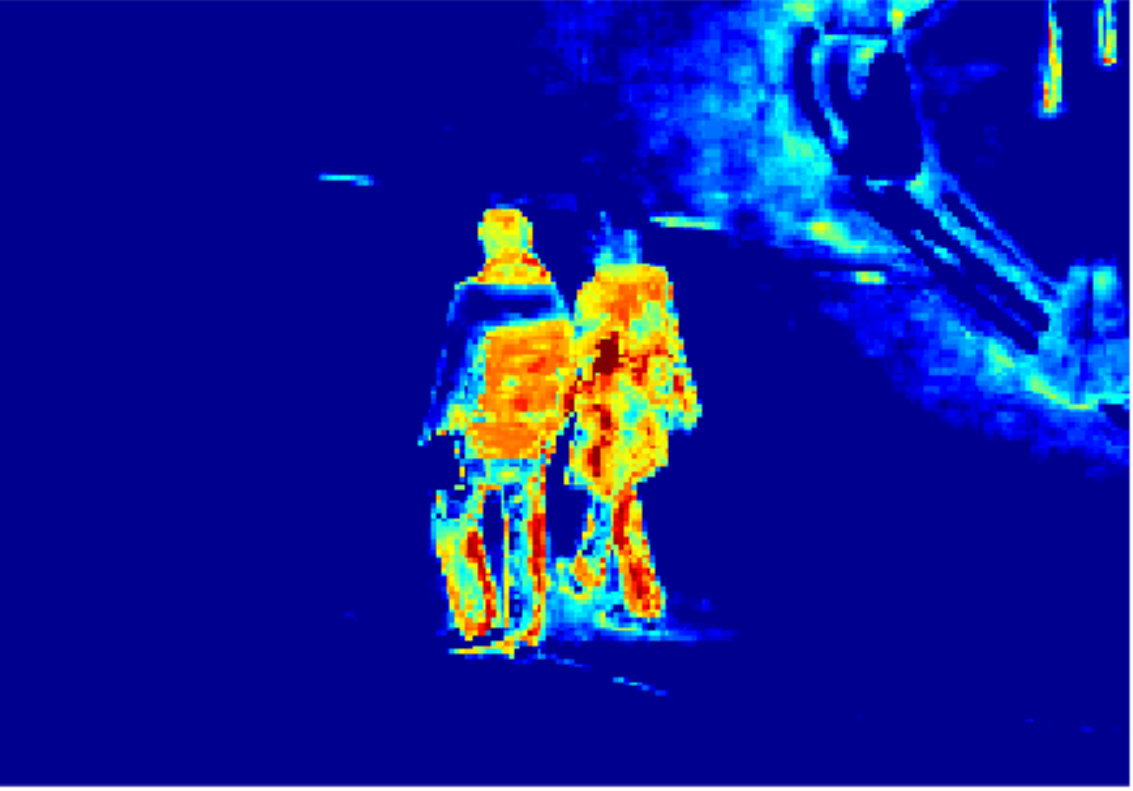}}%
	\subfigure[]{\label{fig:fig2:f}
		\includegraphics[height=1.35cm]{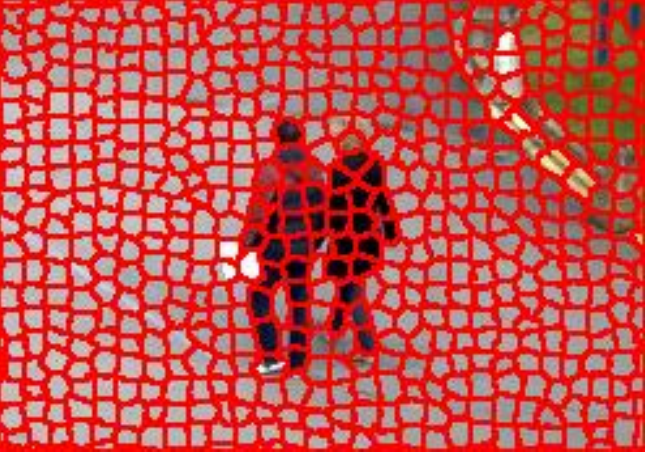}}%
	\subfigure[]{\label{fig:fig2:g}
		\includegraphics[height=1.35cm]{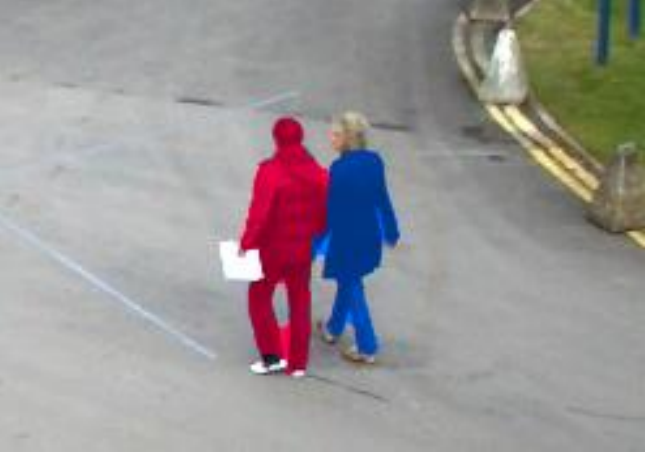}}%
	\caption{An illustration of target/background confidence maps and segmentation results. (a) A new frame (part of the frame is shown for clarity). (b) Background confidence map. Red represents higher confidence value while blue represents lower value. (c) and (d) show confidence maps for the target on the left and the target on the right respectively. (e) Superpixels in the part of the frame. (f) The final segmentation results after applying CRF to the superpixel based spatio-temporal graph. Red and blue masks represent foreground pixels for the two targets respectively.}
	\label{fig:fig2}
\end{figure}

We infer the segmentation of target $i$ in its first frame automatically from its initial box $\bar{y}_i$. GrabCut algorithm \cite{rother2004grabcut} is applied to target $i$'s small surrounding region by initializing pixels within box $\bar{y}_i$ as foreground while pixels outside box $\bar{y}_i$ as background. GrabCut starts from this initial segmentation and iteratively refines foreground/background boundary. Then based on the foreground pixels obtained by GrabCut, we build a pixel-level foreground GMM model $\mathbf{w}_{fg(i)}$ for target $i$. In addition, a background image, obtained by averaging frames in the video, is used to build a universal background GMM model $\mathbf{w}_{bg}$. $CIELAB$ color space is used. A foreground confidence map $S_{fg(i)}$ for target $i$ and a background confidence map $S_{bg}$ are computed by applying $\mathbf{w}_{fg(i)}$ and $\mathbf{w}_{bg}$ to every pixel in a new frame respectively. An example is shown in Fig. \ref{fig:fig2}.

Given $N$ targets in the scene, the goal of segmentation is to assign one of $N+1$ labels ($N$ targets or background) to every pixel. The segmentation problem in upcoming frames is solved by multi-label CRF. Since superpixels naturally preserve the boundary of objects and are computationally efficient for processing, we build a superpixel based spatio-temporal graph. Simple Linear Iterative Clustering (SLIC) \cite{achanta2012slic} is employed to generate $N_{sp}$ superpixels in every frame. There are two types of edges in the graph: spatial edges, $\varepsilon_S$, and temporal edges, $\varepsilon_T$. Spatial edges connect all neighboring superpixels in a frame. Two superpixels $s_k$ and $s_l$ are considered as spatial neighbors if they share an edge in image space. Temporal edges connect all neighboring superpixels in two consecutive frames. Superpixels $s_k$ and $s_l$ are considered as temporal neighbors if at least $1/3$ of the pixels in $s_k$ move to $s_l$ in the next frame as predicted by optical flow. Temporal edges help preserve segmentation consistency across frames.

With the spatio-temporal graph, the multi-label Conditional Random Field (CRF) energy function is defined as
\begin{equation}
\label{eq:energy_seg}
\begin{split}
E_{seg}(Z)=\sum_{s_k}Q(s_k,z_{s_k})+\beta_1\sum_{(s_k,s_l)\in\varepsilon_S}D(s_k,s_l) +\beta_2\sum_{(s_k,s_l)\in\varepsilon_T}D(s_k,s_l),
\end{split}
\end{equation}
where $Z$ denotes the target/background labeling of all superpixels in a segment of video. $z_{s_k}$ is the labeling of superpixel $s_k$. $z_{s_k}=i$ if $s_k$ is labelled as target $i$ and $z_{s_k}=0$ if $s_k$ is labelled as background. The energy function is optimized using graph cuts with $\alpha$-expansion \cite{boykov2001fast}.

The unary term $Q(s_k,z_{s_k})$ in Eq. \ref{eq:energy_seg} is the cost of labeling superpixel $s_k$:
\begin{equation}
Q(s_k,z_{s_k})=\left\{ \begin{array}{rl}
-log(S_{fg(i)}(s_k)), &\mbox{if $z_{s_k}=i$} \\
-log(S_{bg}(s_k)), &\mbox{if $z_{s_k}=0$}\\
\end{array} \right.
\end{equation}
Here $S_{fg(i)}(s_k)$ represents the probability that superpixel $s_k$ belongs to target $i$. It is computed as the average confidence value of $S_{fg(i)}$ over all pixels in $s_k$. $S_{bg}(s_k)$ denotes the probability that superpixel $s_k$ belongs to the background.

The pairwise terms in Eq. \ref{eq:energy_seg} incorporate pairwise constraints by combining color similarity and the mean flow direction similarity between two neighboring superpixels. The pairwise potential $D(s_k,s_l)$ between two spatial/temporal neighboring superpixels $s_k$ and $s_l$ is defined as
\begin{equation}
\centering
\begin{split}
& D(s_k,s_l) =\mathbf{1}(z_{s_k}\neq z_{s_l})\cdot D_c(s_k,s_l)\cdot D_f(s_k,s_l), \\
D_c(s_k,s_l) = & \frac{1}{1+\lVert LAB(s_k)-LAB(s_l)\rVert}, \quad D_f(s_k,s_l) =\frac{V_{s_k}V_{s_l}}{\left \| V_{s_k} \right \|\left \| V_{s_l} \right \|},
\end{split}
\end{equation}
where $\mathbf{1}(\cdot)$ is the one-zero indicator function. $LAB(s_k)$ is the average LAB color of superpixel $s_k$ and $D_c(s_k,s_l)$ defines the color similarity between superpixels $s_k$ and $s_l$. $V_{s_k}$ denotes the mean flow of superpixel $s_k$ and $D_f(s_k,s_l)$ is the direction similarity between the mean flows of superpixels $s_k$ and $s_l$.

\section{Dual Decomposition}
The two tasks - multiple target tracking and target segmentation, discussed in Sec. \ref{sec:subproblems} are highly related. To take advantage of synergies between them, dual decomposition is employed to couple these two tasks. We aim at minimizing the following energy function:
\begin{equation}
\label{eq:energy}
\min_{Y,Z}E(Y,Z) = \min_{Y,Z}(E_{track}(Y)+E_{couple}(Y,Z)+E_{seg}(Z)),
\end{equation}
where $E_{track}(Y)$ and $E_{seg}(Z)$ are defined as in Eq. \ref{eq:network_flow} and Eq. \ref{eq:energy_seg} respectively. $Y$ denotes the set of bounding boxes found by the tracking procedure in Sec. \ref{sec:track} and $Z$ denotes target/background segmentation obtained in Sec. \ref{sec:seg}. The coupling term contains both bounding boxes and segmentation information:
\begin{equation}
\begin{split}
E_{couple}(Y,Z) = \sum_{i,k}(\mathbf{1}(k\in y_i,z_k\neq i)\theta_{y_i}^k + \mathbf{1}(k\notin y_i,z_k=i)\varphi_{y_i}^k).
\end{split}
\end{equation}
This energy introduces penalties for background labels inside target bounding boxes as well as foreground labels outside target bounding boxes. $i$ denotes a target and $k$ denotes a pixel. The first term penalizes pixels that are not labelled as target $i$, but are in target $i$'s tracking boxes. $y_i$ denotes the bounding box for target $i$, and $\mathbf{1}(k\in y_i,z_k\neq i)$ represents pixels in $y_i$ which are not labelled as target $i$. Since a target's bounding box is highly likely to include some non-target pixels near the border of box, but not at the center of box, the resulting penalty is weighted by a human shape prior $\theta_{y_i}$. Thus, background pixels at the center of box induce higher penalty while those close to the border of box result in lower penalty. The same human shape prior $\theta$ is used as in \cite{milan2015joint}. The second term penalizes pixels that are labelled as target $i$ but are outside target $i$'s boxes. $\mathbf{1}(k\notin y_i,z_k=i)$ represents pixels outside $y_i$ which are labelled as target $i$. The corresponding penalty is weighted by $\varphi_{y_i}$, which has a zero weight within $y_i$ and uniform non-zero weight outside $y_i$.

By introducing an equality constraint, Eq. \ref{eq:energy} can be rewritten as
\begin{equation}
\label{eq:energy2}
\begin{split}
\hspace{0cm} \min_{Y^0,Y^1,Z}E(Y^0,Y^1,Z) = & \min_{Y^0,Y^1,Z}(E_{track}(Y^0) + E_{couple}(Y^1,Z) + E_{seg}(Z)) \\
& s.t. \quad Y^0=Y^1.
\end{split}
\end{equation}

Now, the energy function is separable. We form the Lagrangian dual form of the above problem by introducing Lagrange multipliers $\lambda$
\begin{equation}
\label{eq:energy_dual}
\begin{split}
L(\lambda) = & \min_{Y^0,Y^1,Z}(E_{track}(Y^0) + E_{couple}(Y^1,Z)+E_{seg}(Z) + \lambda(Y^0-Y^1)), \\
= & \min_{Y^0}(E_{track}(Y^0)+\lambda Y^0) + \min_{Y^1,Z}(E_{couple}(Y^1,Z) + E_{seg}(Z)-\lambda Y^1). \\
\end{split}
\end{equation}
Here $\lambda$ has the same dimension as $Y^0$ and $Y^1$.

Eq. \ref{eq:energy_dual} can be further decomposed into two independent subproblems:
\begin{align}
g(\lambda) & = \min_{Y^0}(E_{track}(Y^0)+\lambda Y^0),\label{eq:energy_g} \\
h(\lambda) & = \min_{Y^1,Z}(E_{couple}(Y^1,Z)+E_{seg}(Z)-\lambda Y^1).\label{eq:energy_h}
\end{align}
The first subproblem (Eq. \ref{eq:energy_g}) is equivalent to a set of network flow problems, thus $g(\lambda)$ can be solved efficiently using dynamic programming. The second subproblem (Eq. \ref{eq:energy_h}) involves both tracking boxes and segmentation. When $Y^1$ is fixed, $E_{couple}(Y^1,Z)$ becomes a unary term on $Z$, thus $h(\lambda)$ can be solved by graph-cut. When $Z$ is fixed, $h(\lambda)$ can be optimized by evaluating all candidate boxes. So a two-step procedure is employed to optimize $h(\lambda)$.

We use a sub-gradient method to optimize the Lagrangian dual problem. The algorithm works by repeating the following steps:
\begin{enumerate}
	\item Get $Y^0$ by solving the tracking subproblem $g(\lambda)$ (Eq. \ref{eq:energy_g}).
	\item Get $Y^1$ by solving the segmentation subproblem $h(\lambda)$ (Eq. \ref{eq:energy_h}).
	\item Stop if $Y^0=Y^1$.
	\item Otherwise, update dual variable $\lambda$ by $\lambda \leftarrow \lambda +\alpha_t (Y^0-Y^1)$, where $\alpha_t$ is the step size in iteration $t$ and is computed as $\alpha_t=1/(10+t)$.
\end{enumerate}

In each iteration, we check the consistency between solutions of the two subproblems. The dual variable $\lambda$ changes based on the inconsistent parts among $Y^0$ and $Y^1$, thus adjusting $Y^0$ and $Y^1$ accordingly to make them to be more and more consistent. Suppose in some iteration, box $y_i$ is selected for target $i$ by the tracking subproblem, but the segmentation subproblem selects another box. Then the corresponding element in $\lambda$ will increase such that the penalty of selection of  $y_i$ by the tracking subproblem would increase and the penalty of selection of $y_i$ by the segmentation subproblem would decrease. When $Y^0$ and $Y^1$ achieve agreement, $\lambda$  will not change and the optimal solution is found.

In traditional tracking-by-detection approaches, non-maximum-suppression is usually applied to human detection results before the data association step. This is to prevent multiple tracks from selecting the same target. Since our algorithm considers dense and overlapping candidate boxes, we incorporate a spatial constraint in tracking in order to overcome the same issue. Let $\gamma$, which has the same dimension as $Y^0$, defines the cost in network flow problem induced by the spatial constraint. The tracking subproblem Eq. \ref{eq:energy_g} now becomes
\begin{equation}
g(\lambda)=\min_{Y^0}(E_{track}(Y^0)+\lambda Y^0+\gamma Y^0),
\end{equation}
where $\gamma$ is first initialized as an all-zero matrix and then updated in each iteration according to the tracking and segmentation results in the last iteration. Assume that from the tracking results in iteration $t-1$, a box $y_i$ is selected for target $i$. If the distance between $y_i$ and any box in other tracks is smaller than $\delta$, and no pixel in $y_i$ is labelled as target $i$ from the segmentation results, there is a large chance that box $y_i$ does not correspond to target $i$ well. Thus the corresponding element in $\gamma$ is updated as $\gamma \leftarrow \gamma +\alpha_t$. In this way, box $y_i$ will introduce larger penalty and be less likely selected in tracking in iteration $t$. Note that the segmentation results are considered along with the tracking results, so the spatial constraint introduces penalty only if a box is too close to another box and is not supported by the segmentation results. This happens when two targets are close to each other, and the track of one target incorrectly jumps to the other target. On the contrary, when one target is occluded by another target, even though their boxes are close, they both have supporting pixels from the segmentation results, therefore no spatial constraint should apply.

\begin{figure}[!b]
	\centering
	\subfigure[]{\label{fig:fig5:a}
		\includegraphics[width=0.31\textwidth]{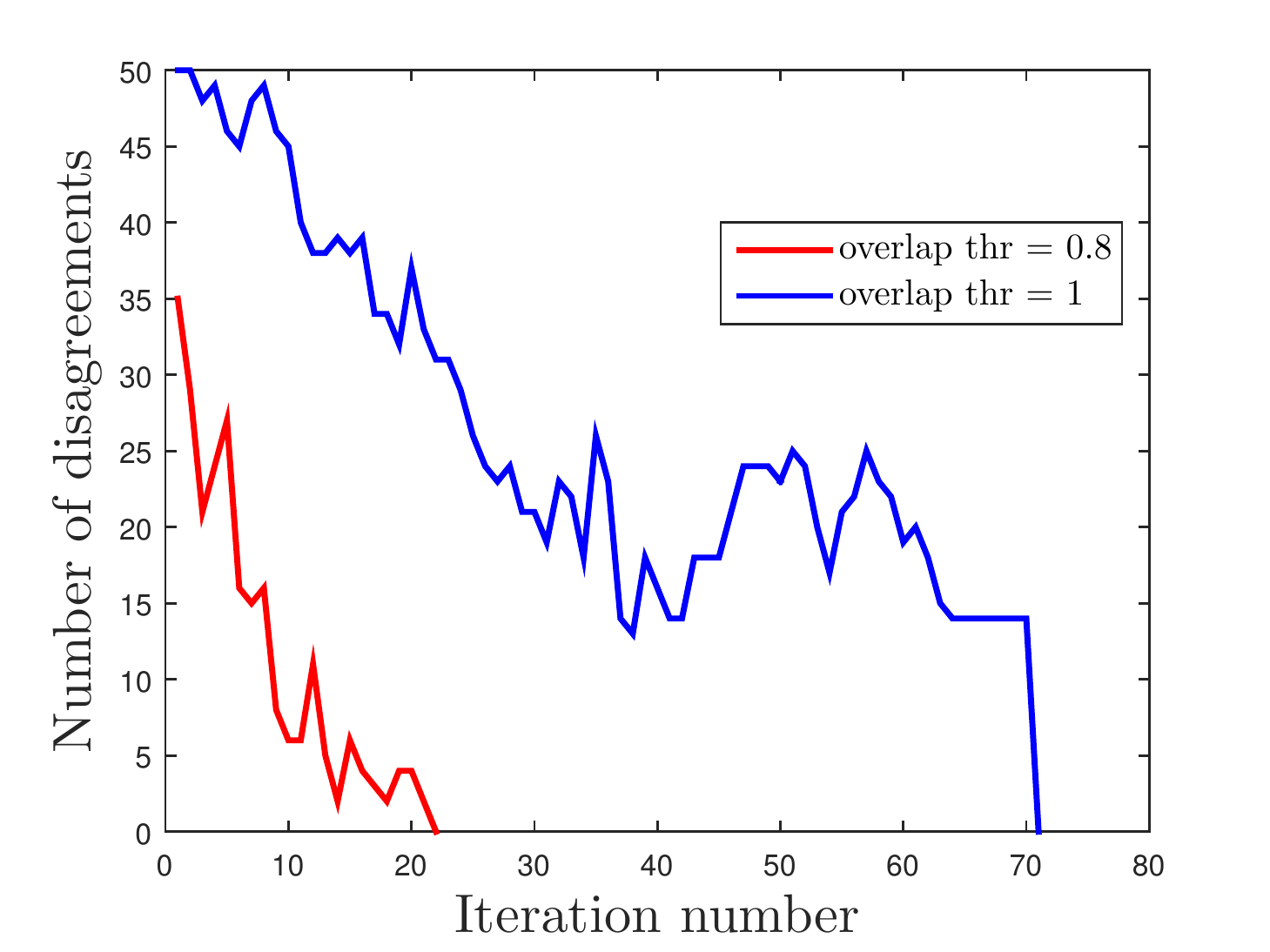}}%
	\subfigure[]{\label{fig:fig5:b}
		\includegraphics[width=0.31\textwidth]{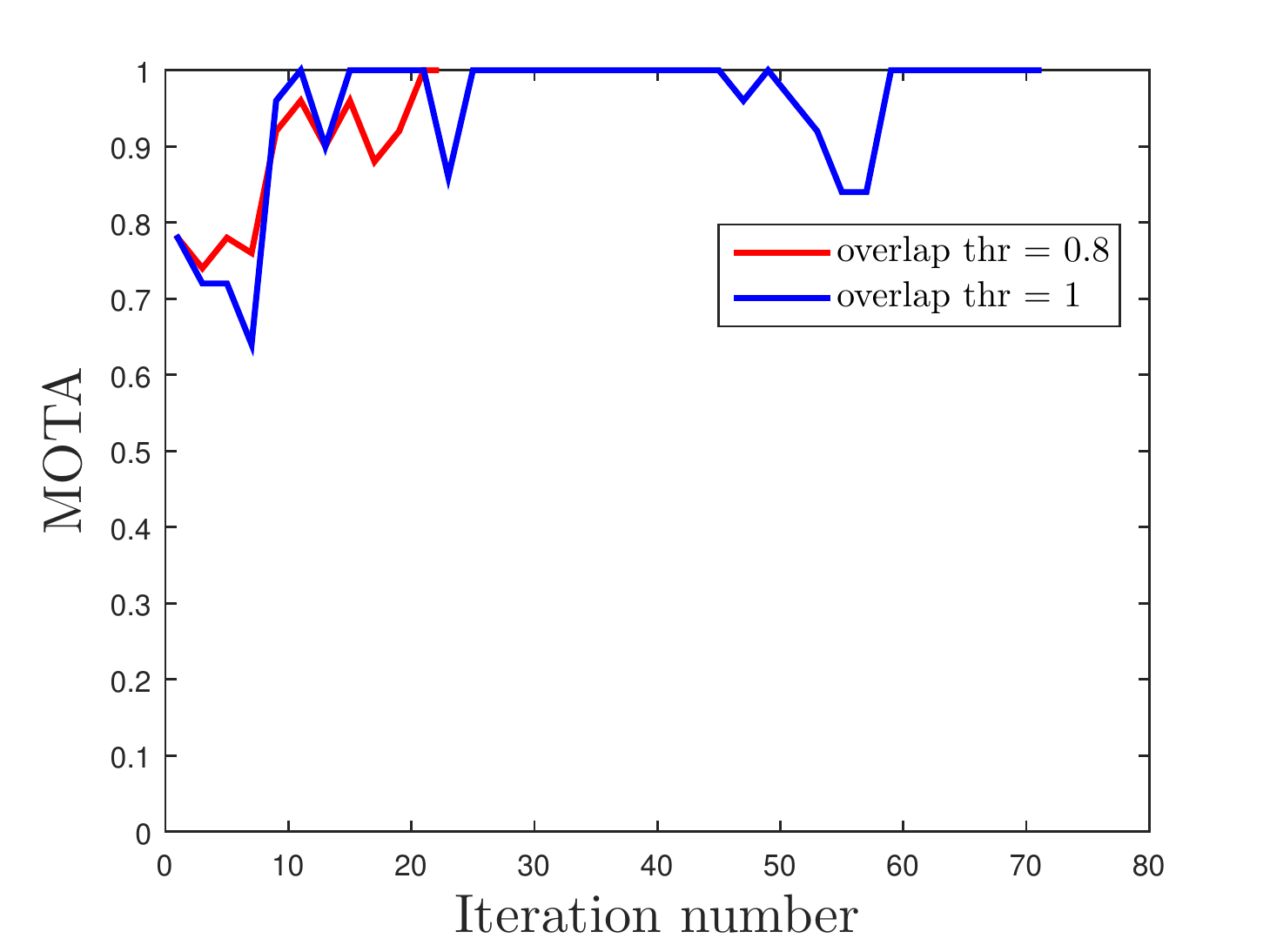}}%
	\subfigure[]{\label{fig:fig5:c}
		\includegraphics[width=0.31\textwidth]{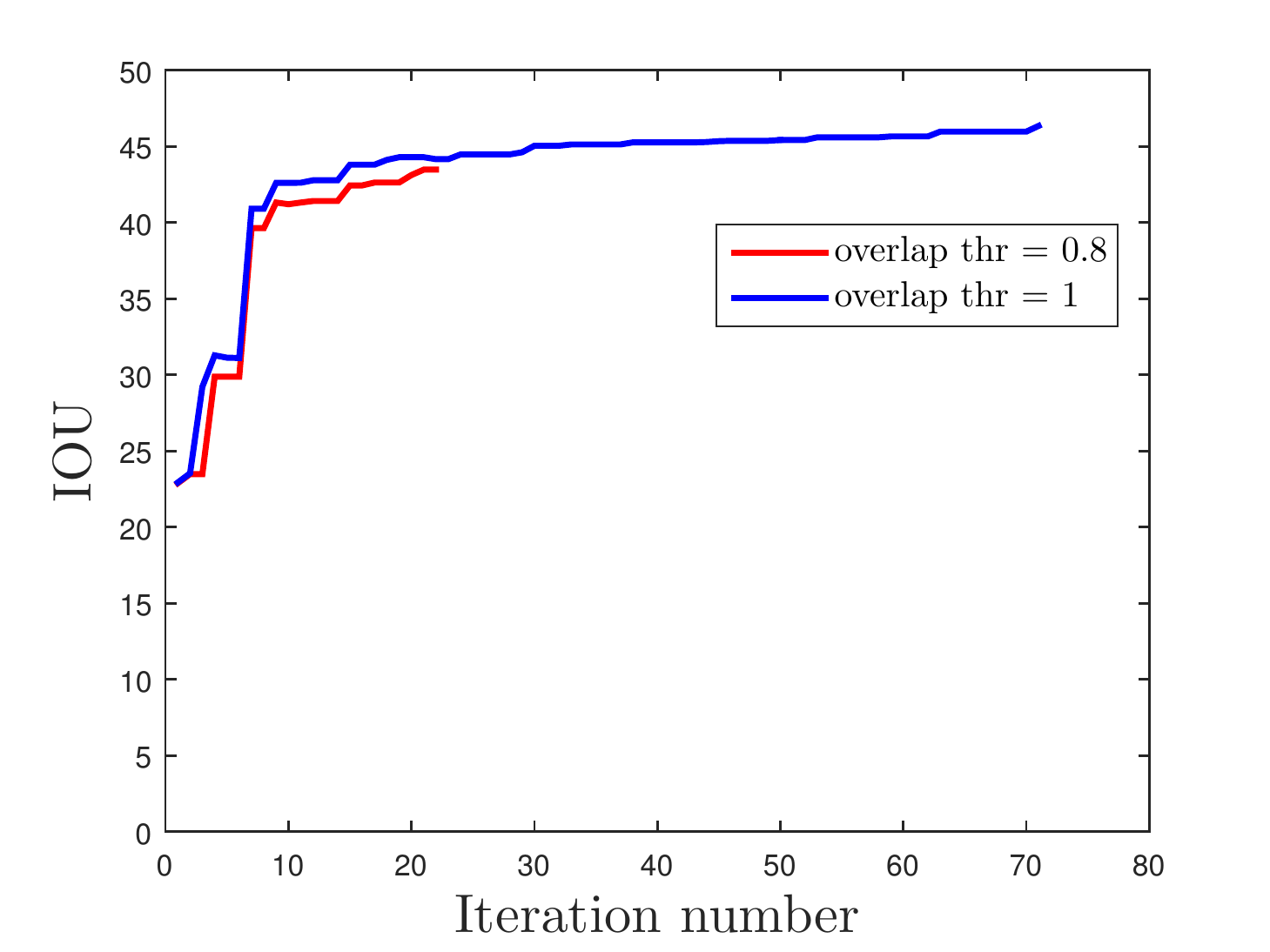}}
	\caption{The curves are generated based on a $10$-frame segment in TUD-Crossing with $5$ persons in the scene. (a) The number of disagreements between tracking and segmentation solutions drops over iterations. The algorithm converges when the two solutions are consistent. (b) The MOTA increases over iterations and reaches the best value at convergence. (c) The IOU (metric detailed in Sec. \ref{sec:exp_seg}) increases over iterations. Since the segmentation annotations are available in every $10$ frame, IOU is evaluated on the one frame in the $10$-frame segment which has segmentation annotations.}
	\label{fig:fig5}
\end{figure}

Due to the dense and overlapping candidate boxes used in our approach, we observe it is not necessary to have $Y^0$ and $Y^1$ to be exactly the same for convergence. In most cases, the results in early iterations are already good enough, though some boxes found by the two subproblems may shift a little. In our experiments, boxes returned by the two subproblems are considered consistent if their overlap is larger than $0.8$ and the corresponding element in $\lambda$ will not be updated. This greatly reduces the number of iterations to solve the Lagrangian dual problem, with almost no performance loss. As shown in Fig. \ref{fig:fig5:a}, when overlap threshold of $0.8$ is used, the number of disagreements drops more quickly compared to that of using overlap threshold of $1$. The number of iterations to solve the Lagrangian dual problem is reduced by more than three times. Meanwhile, the performance remains almost the same as illustrated in Fig. \ref{fig:fig5:b} and \ref{fig:fig5:c}.

Coupling tracking and segmentation lead to both better tracking and better segmentation results as demonstrated in experiments. It can also be observed in Fig. \ref{fig:fig5} that the MOTA and IOU are increasing over iterations. On one hand, the object tracks provide strong high-level guidance for target/background segmentation. On the other hand, segmentation helps resolve typical difficulties in multiple target tracking. First, in traditional tracking-by-detection approach, the tracking results highly depend on the detection performance. Miss-detections are common especially when there is occlusion. So special scheme, such as dummy nodes in network flow, needs to be designed in order to handle them. However, our approach does not rely on pre-trained object detector. We assume densely sampled candidate boxes instead of sparse detection boxes, so the tracker is able to infer temporal consistency between frames naturally. In addition, when target gets occluded, its visible part is segmented correctly even though its overall appearance score may be low. The segmentation result would guide tracker to find correct box for the target. Second, the segmentation result provides more information about target location and target identity. Therefore, it helps tracker avoid drifting and ID-switch.

\section{Experiments}
\subsection{Datasets and Experimental Setup}
We test our proposed algorithm on three publicly available sequences: PETS-S2L1 \cite{ferryman2010pets2010}, TUD-Crossing \cite{andriluka2008people} and TUD-Stadtmitte \cite{andriluka2010monocular}. The sequences involve different camera angles, frequent occlusions and dynamic target behaviors, posing difficulty in individual tracking and segmentation.

In all experiments, the number of components in each foreground GMM and that in the universal background GMM are $10$ and $50$ respectively. The weight for spatial and temporal neighboring superpixels' distance are set as $\beta_1=1$ and $\beta_2=5$ in Eq. \ref{eq:energy_seg}. In TUD-Crossing and TUD-Stadtmitte the number of superpixels in each frame $N_{sp}=2000$, while in PETS-S2L1 $N_{sp}=5000$, due to the fact that persons in the last sequence are much smaller than those in the first two sequences. The costs to start and end a track $c_n^s$ and $c_n^t$ in Eq. \ref{eq:network_flow} are both set to $10$. The distance threshold $\delta$ for spatial constraint is set to $15$ pixels.

Similar to \cite{zhang2013structure,breitenstein2011online,dehghan2015target}, we use the first bounding box of each target from the annotation to learn discriminative appearance model and start tracking in the following frames. The algorithm runs in online manner and dual decomposition is applied to every 10 frames segment of a video. When a target is close to the scene border and its velocity is towards outside of the scene, that target is treated as exiting the scene and the algorithm stops tracking that target. In this way, our approach is able to handle a variable number of targets in the scene. The experiment in this setting is denoted as ``Ours" in evaluation. Besides manual initialization, we also conduct a set of experiments using human detector \cite{felzenszwalb2010object} to initialize targets, denoted as ``Ours - Auto", which makes our system fully automatic. A new target is initialized if there are at least five confident detections with high overlap in consecutive frames and these detections do not correspond to any existing tracks. Moreover, to demonstrate the benefits of combining segmentation and tracking, two baseline approaches are tested. ``Ours - Seg" represents a baseline with pure segmentation while ``Ours - track" represents a baseline with only the tracking part without segmentation incorporated.

\subsection{Segmentation}\label{sec:exp_seg}
Our approach is able to track multiple targets with pixel-level target/background labeling. In order to evaluate the segmentation performance, we use the segmentation annotations for TUD-Crossing from \cite{horbert2011level} and manually annotate pixel-level target masks every $10$ frames in the other sequences. The segmentation annotations will be released to facilitate future research in this area.

For evaluation, the segments are optimally assigned to ground truth masks and multiple segments can be assigned to the same ground truth mask. Identity-based IOU is the average intersection-over-union overlap with target identity information incorporated. Traditional IOU used in video segmentation evaluation \cite{li2013video} computes the mean IOU of foreground regions over all frames. However, it has no notion of target identities. Therefore, in order to better evaluate the segmentation performance for multiple targets, we extend the traditional foreground IOU to \textbf{identity-based IOU}. Identity-based IOU computes the interection-over-union overlap between ground truth mask and segments assigned to it for every target in every frame and then takes the average over all of them. Overall error is the percentage of wrongly labelled pixels while average error computes the percentage of misclassified pixels per ground truth mask. Over-segmentation counts the number of segments merged to cover the ground truth masks.

\begin{table}[t]
	\scriptsize
	\begin{center}
		\begin{tabu} to \textwidth{|X[1.2,c]|X[1.4,c]|X[1.5,c]|X[0.8,c]|X[0.8,c]|X[0.8,c]|}
			\hline
			Dataset & Method & Identity-based IOU & Overall err. & Avg. err. & Over-seg. \\
			\hline
			\multirow{3}{*}{PETS-S2L1}
			& Milan et al. \cite{milan2015joint} & 54.82 & 0.78 & 40.08 & 1.65 \\
			& \textbf{Ours - Seg} & 19.51 & 1.68 & 66.86 & 1 \\
			& \textbf{Ours - Auto} & 72.78 & 0.42 & 19.11 & 1 \\
			& \textbf{Ours} & 73.51 & 0.43 & 17.79 & 1 \\
			\hline
			\multirow{2}{*}{TUD-Crossing}
			& Milan et al. \cite{milan2015joint} & 25.35 & 6.68 & 63.87 & 2.23 \\
			& Horbert et al. \cite{horbert2011level} & 46.50 & 4.13 & 35.88 & 3.23 \\
			& \textbf{Ours - Seg} & 15.64 & 7.96 & 71.85 & 1 \\
			& \textbf{Ours - Auto} & 51.36 & 4.12 & 30.52 & 1 \\
			& \textbf{Ours} & 55.36 & 3.88 & 26.93 & 1 \\
			\hline
			\multirow{3}{*}{TUD-Stadtmitte}
			& Milan et al. \cite{milan2015joint} & 27.33 & 6.10 & 48.59 & 1.09 \\
			& \textbf{Ours - Seg} & 18.87 & 6.85 & 56.48 & 1 \\
			& \textbf{Ours - Auto} & 40.46 & 3.67 & 23.99 & 1 \\
			& \textbf{Ours} & 41.62 & 3.35 & 23.65 & 1 \\
			\hline
		\end{tabu}
	\end{center}
	\caption{\label{tab:tab2}A comparison of segmentation results on PETS-S2L1, TUD-Crossing and TUD-Stadtmitte.}
\end{table}

We compare the above four metrics with \cite{milan2015joint}\footnote{We test the code available on the author's website with default parameters on TUD-Crossing. The results on the other two sequences are obtained from the author. } and \cite{horbert2011level}\footnote{Note that the identity-based IOU of Horbert et al.'s \cite{horbert2011level} results is computed using the segmentation results provided by the author, while the IOU reported in \cite{horbert2011level} is the traditional foreground IOU without notion of target identities.} in Table \ref{tab:tab2}. The proposed approach achieves much higher identity-based IOU and much lower overall error as well as average error compared to previous methods. ``Ours" outperforms ``Ours - Seg" by a large margin, demonstrating that incorporating tracking leads to more accurate segmentation results. In addition, ``Ours - Auto" achieves comparable results as ``Ours". The performance drops a little compared to ``Ours" because when human detections are used to initialize targets, some targets may be initialized later compared to human initialization. Some qualitative results are shown in Fig. \ref{fig:fig4}. Note that targets are segmented and tracked correctly even when being occluded or when they are close to other targets.

\begin{figure}[t]
	\centering
	\subfigure[]{\label{fig:fig6:a}
		\includegraphics[width=0.31\textwidth]{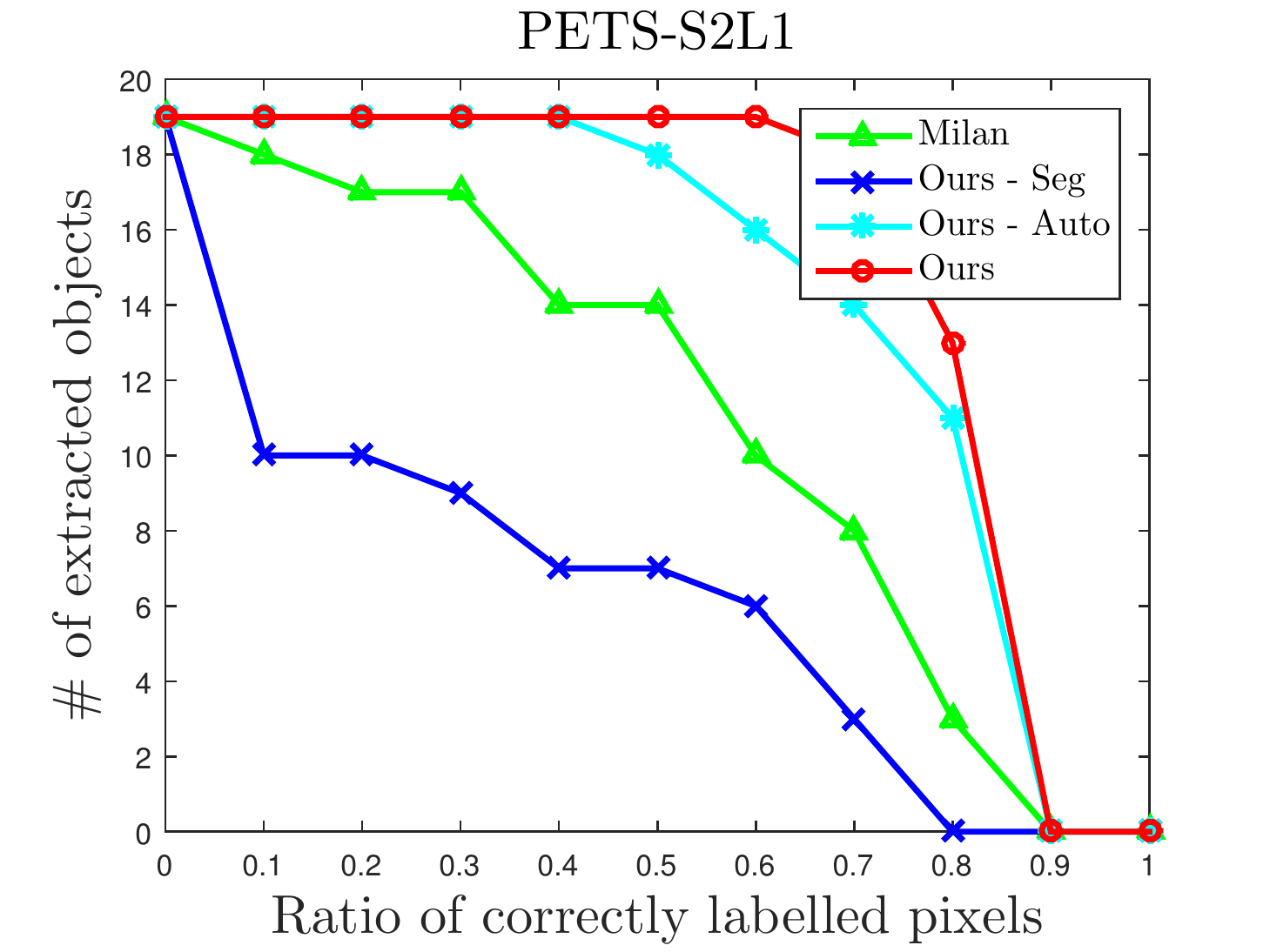}}%
	\subfigure[]{\label{fig:fig6:b}
		\includegraphics[width=0.31\textwidth]{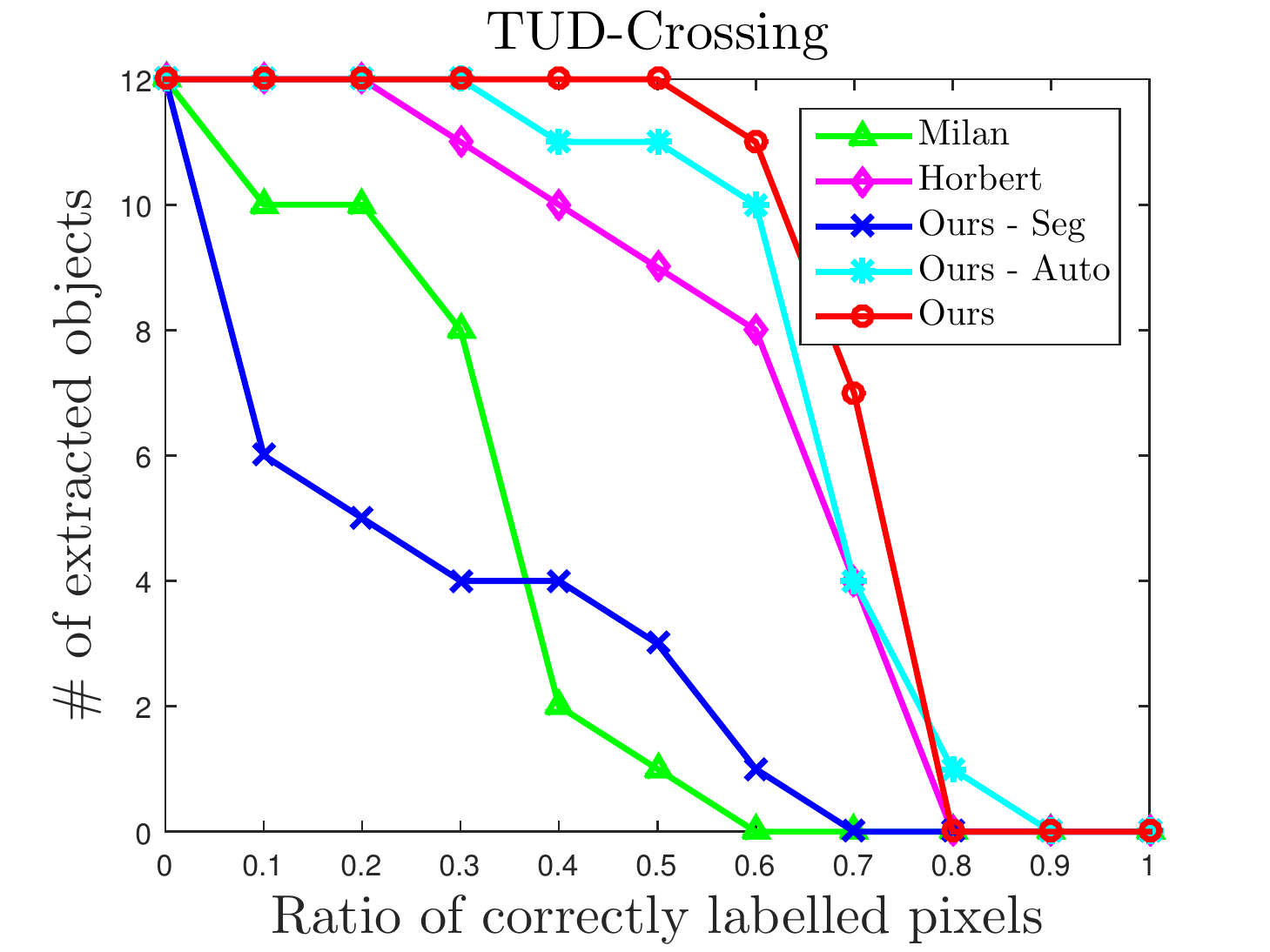}}%
	\subfigure[]{\label{fig:fig6:c}
		\includegraphics[width=0.31\textwidth]{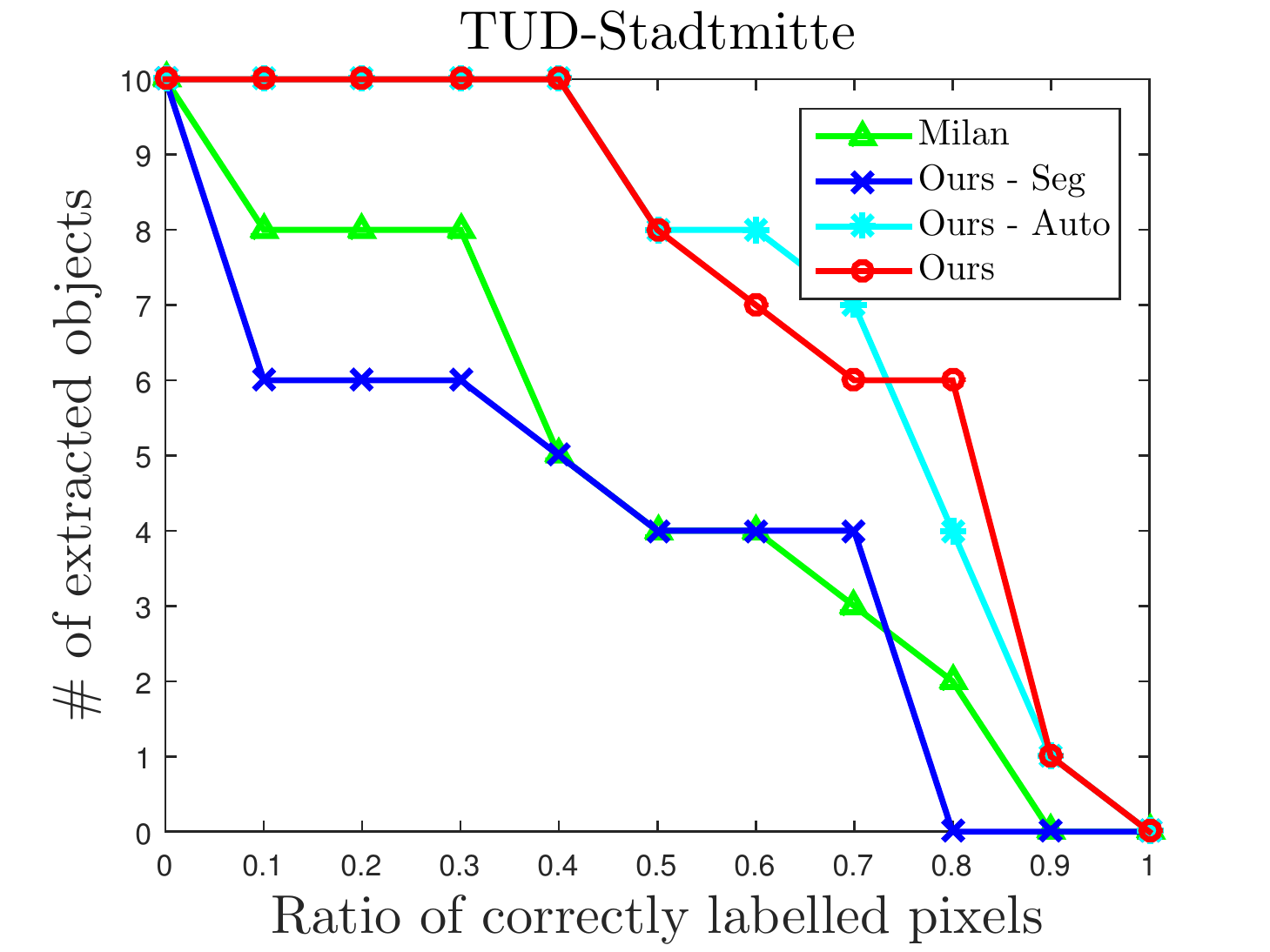}}%
	\caption{The curves show number of extracted objects with varying threshold on ratio of correctly covered region per ground truth mask.}
	\label{fig:fig6}
\end{figure}

\begin{figure}[!b]
	\centering	
	\subfigure{
		\includegraphics[width=0.24\textwidth]{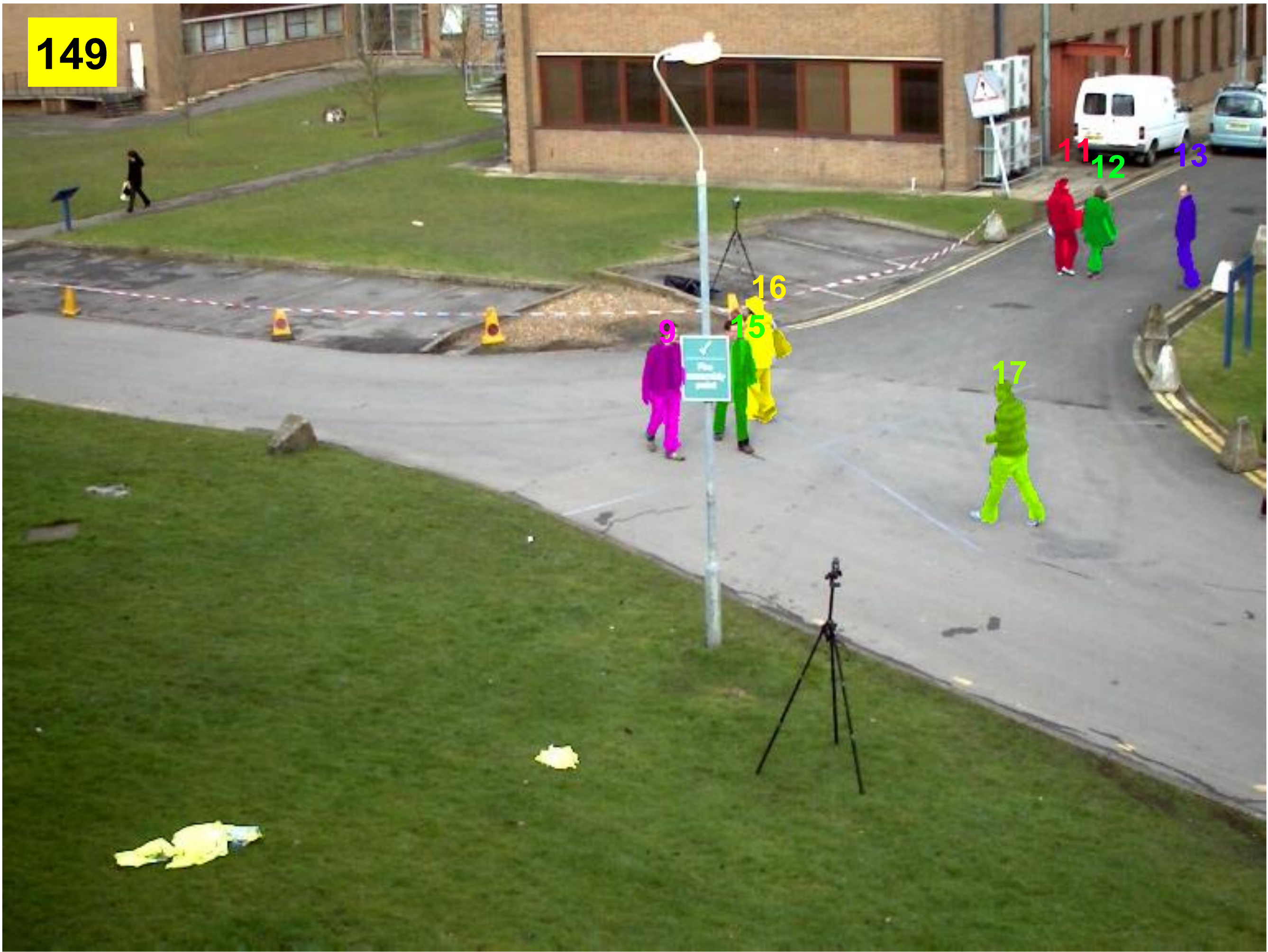}}%
	\subfigure{
		\includegraphics[width=0.24\textwidth]{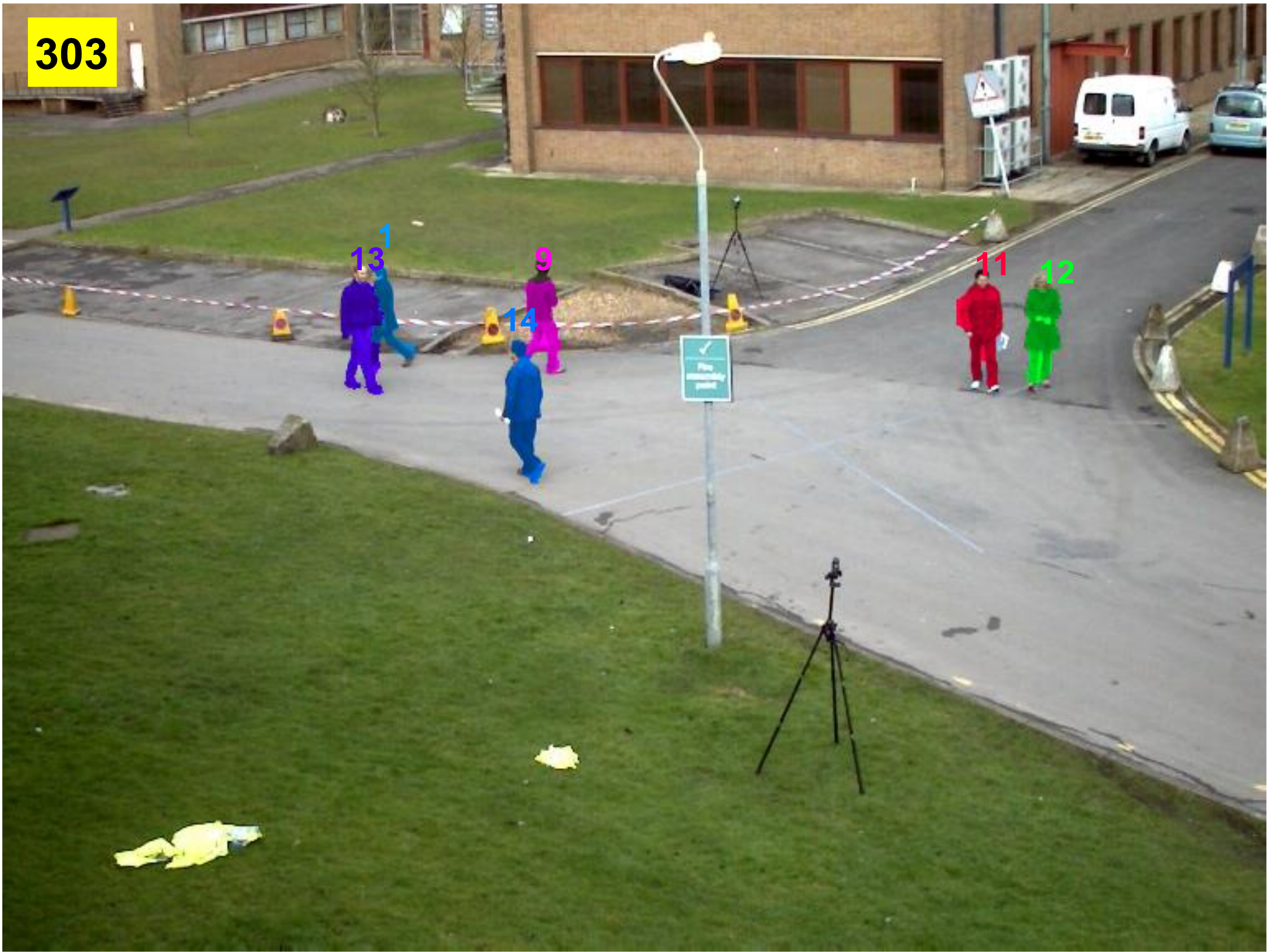}}%
	\subfigure{
		\includegraphics[width=0.24\textwidth]{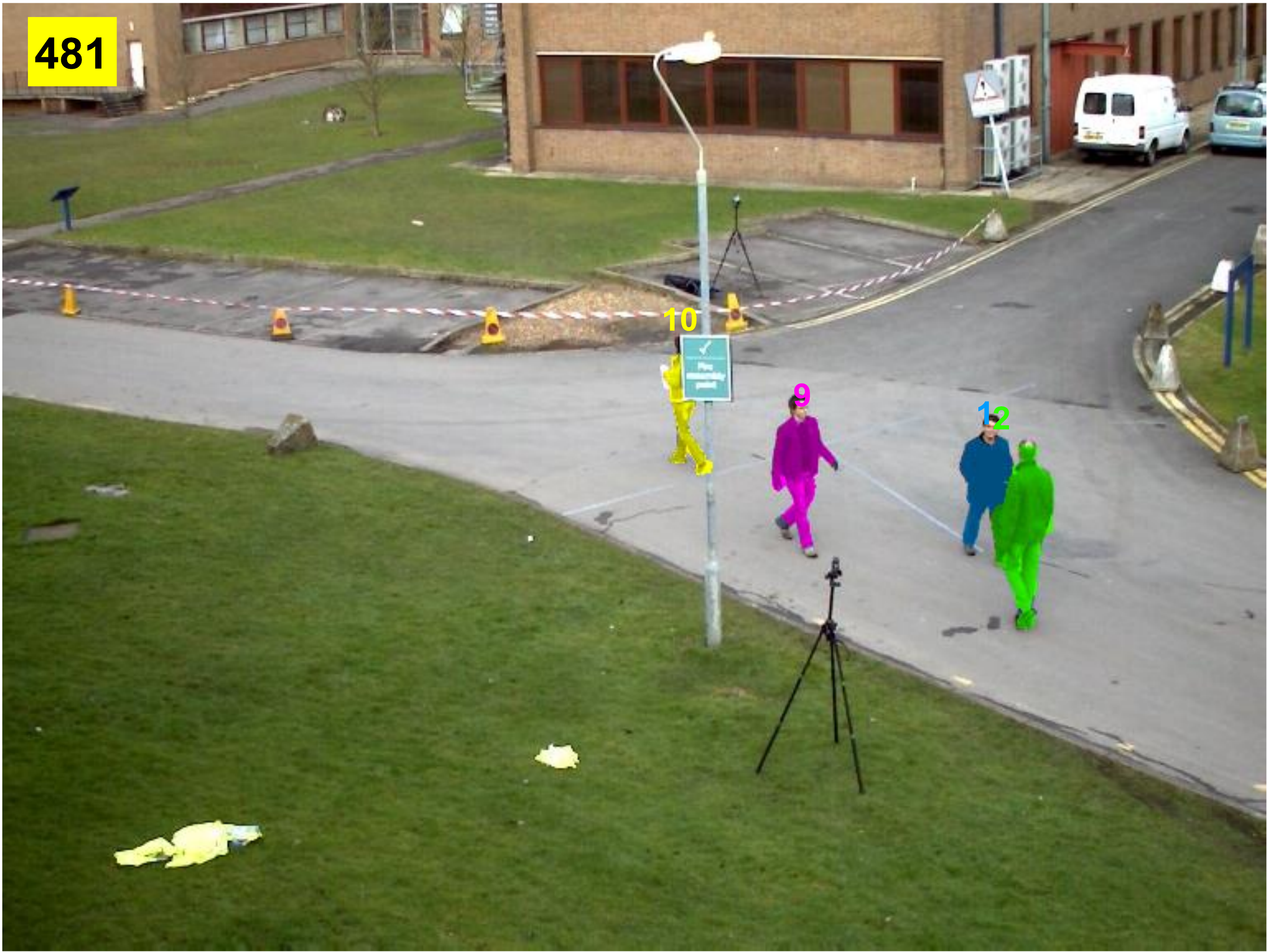}}%
	\subfigure{
		\includegraphics[width=0.24\textwidth]{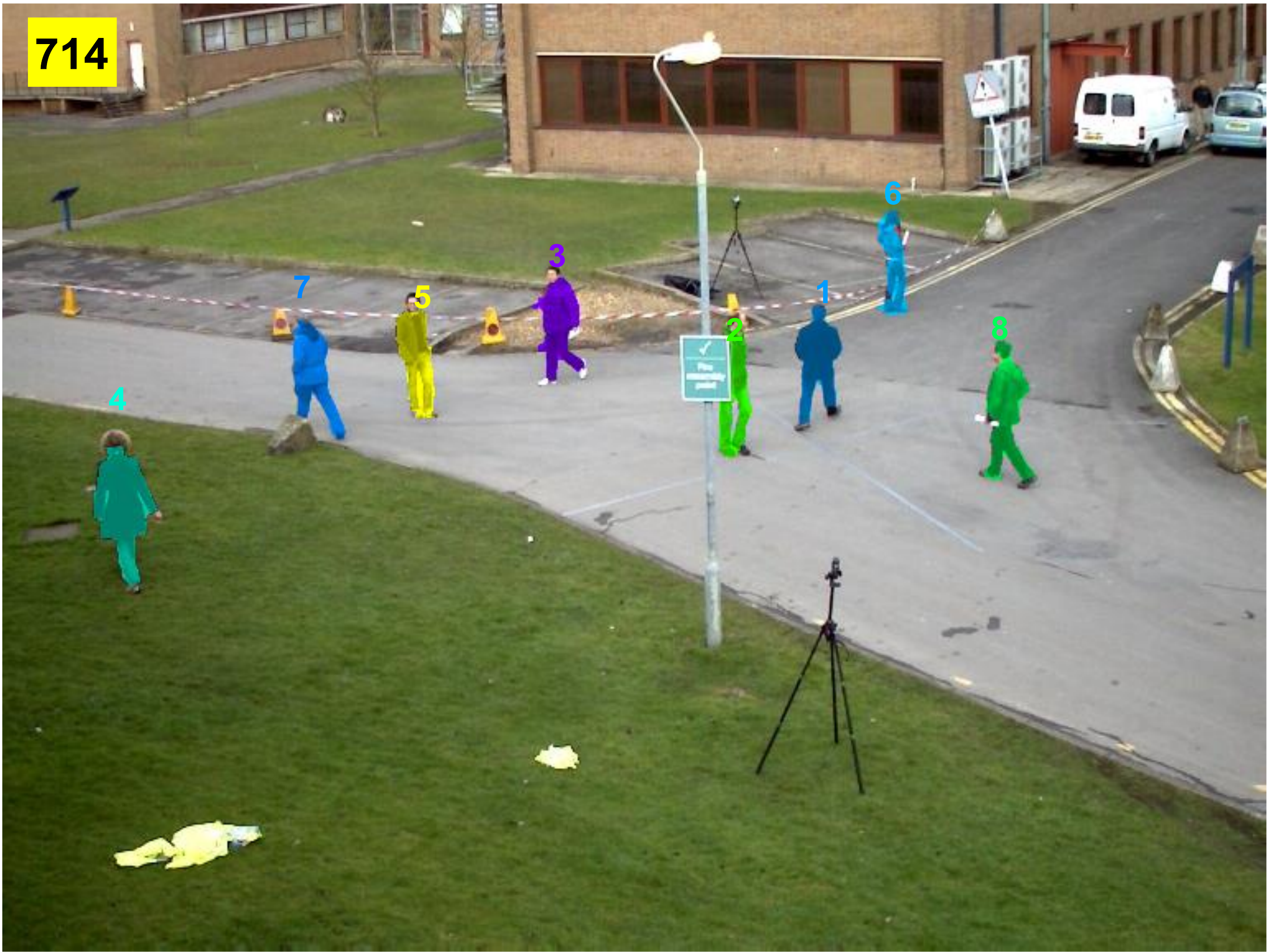}}
	
	\vspace{-0.3cm}	
	\subfigure{
		\includegraphics[width=0.24\textwidth]{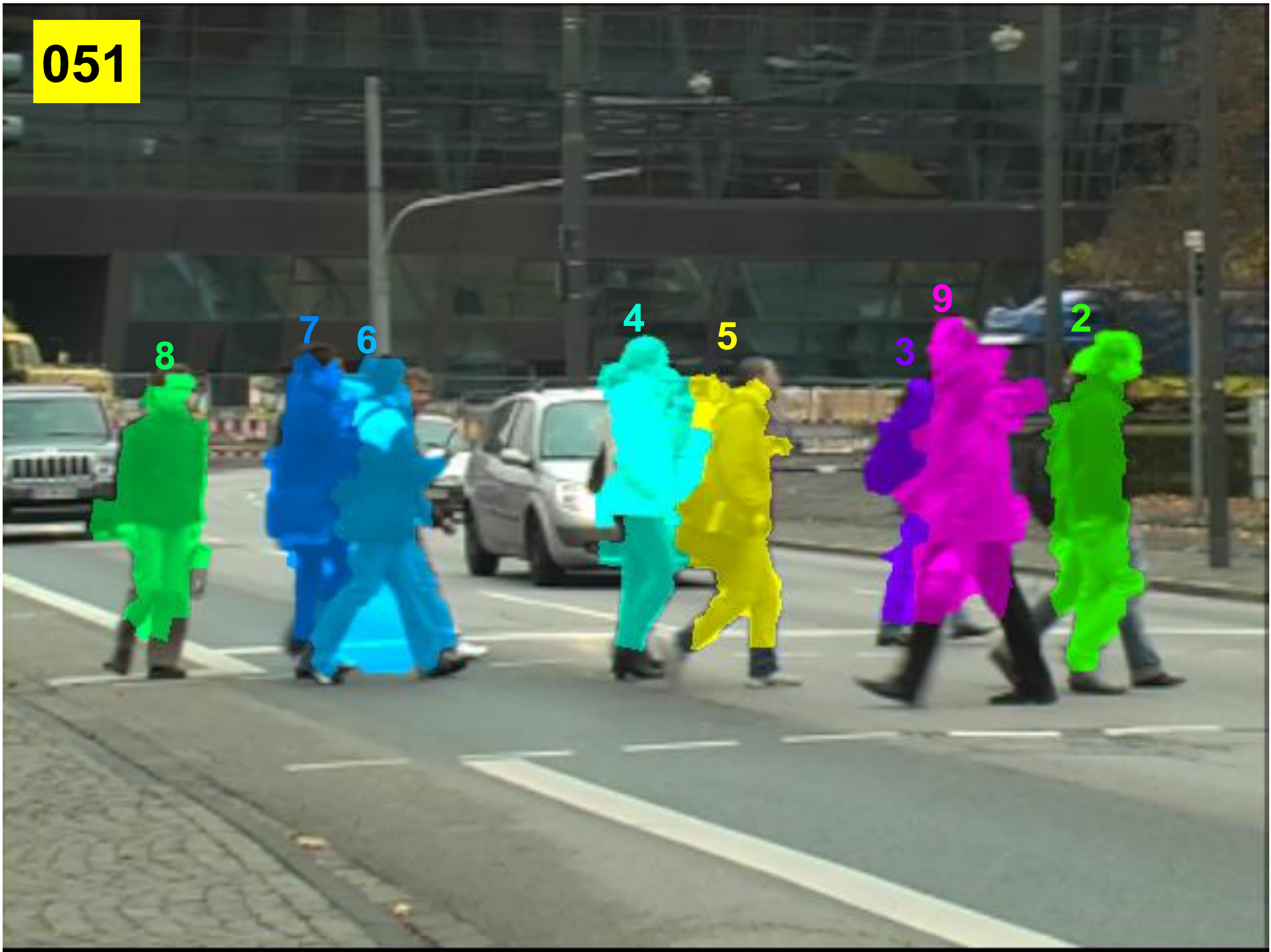}}%
	\subfigure{
		\includegraphics[width=0.24\textwidth]{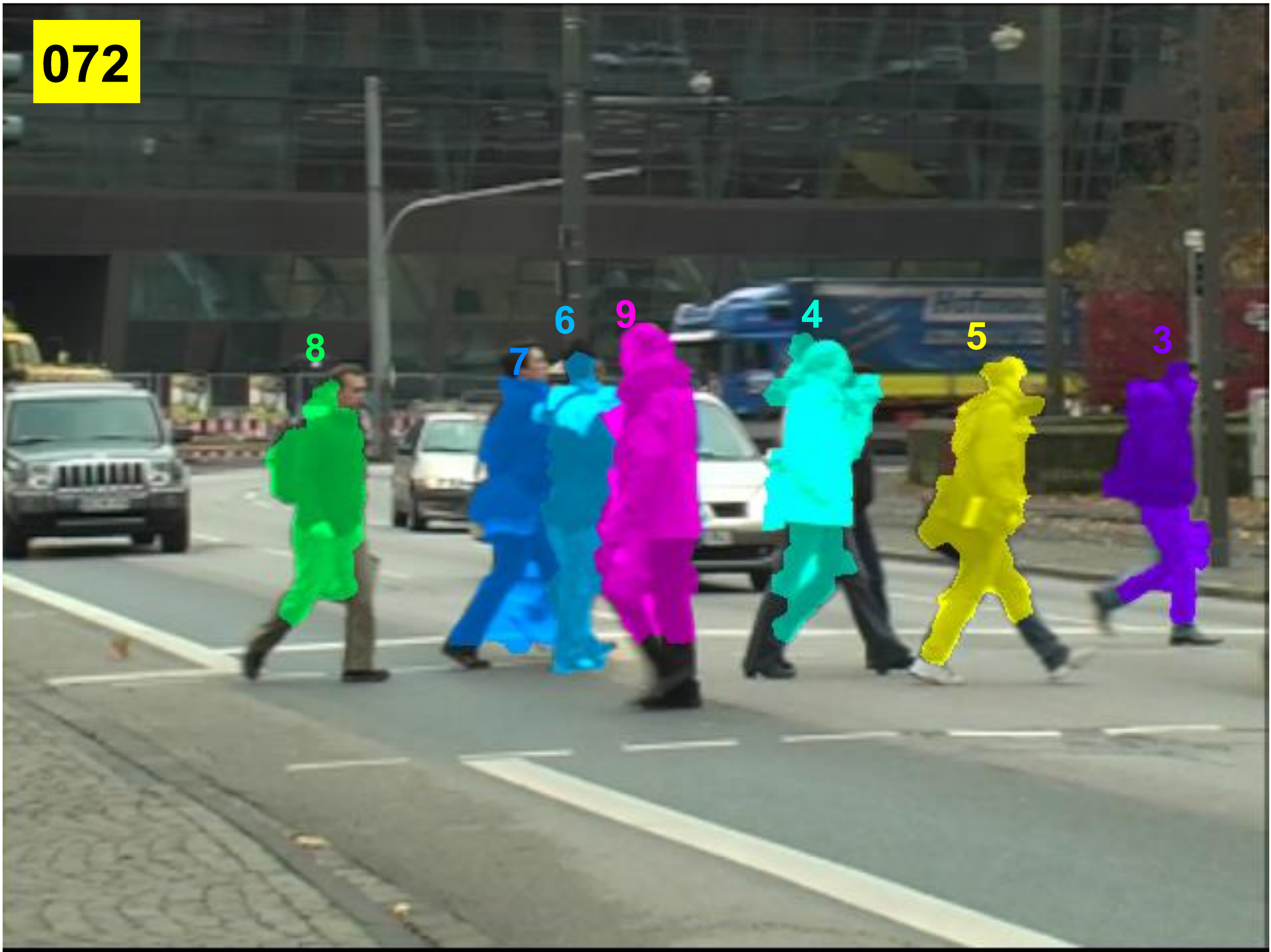}}%
	\subfigure{
		\includegraphics[width=0.24\textwidth]{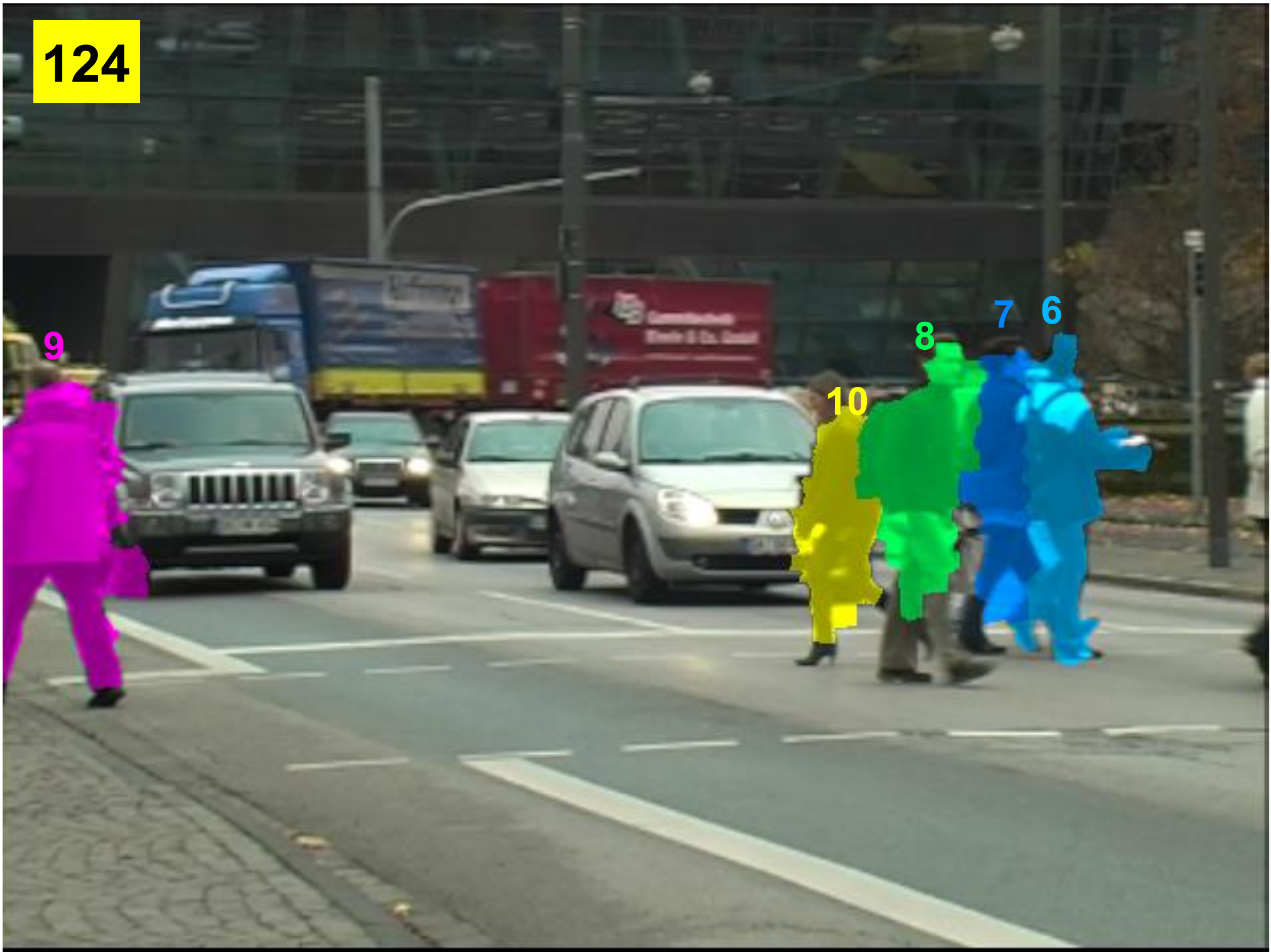}}%
	\subfigure{
		\includegraphics[width=0.24\textwidth]{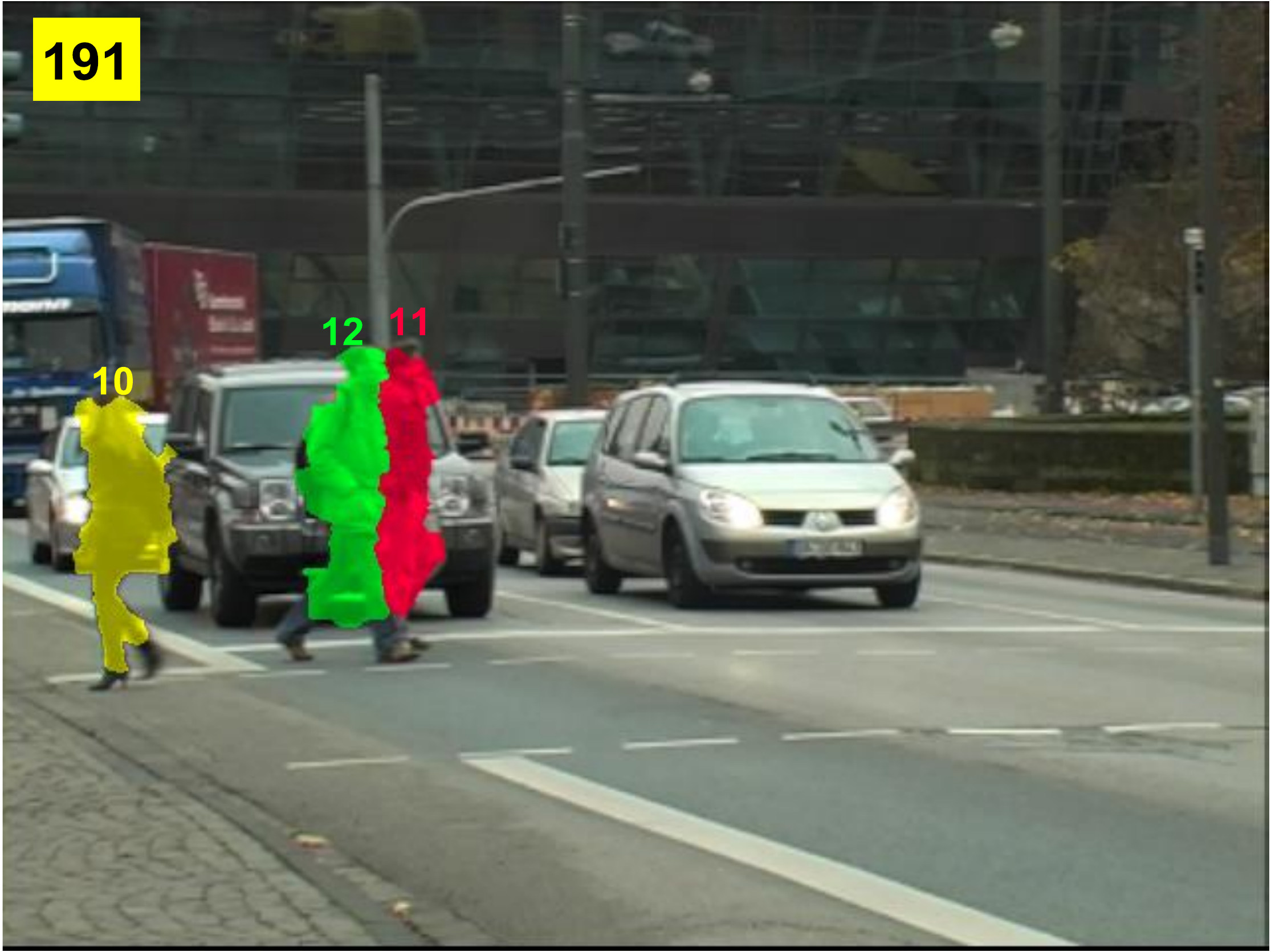}}
	
	\vspace{-0.3cm}
	\subfigure{
		\includegraphics[width=0.24\textwidth]{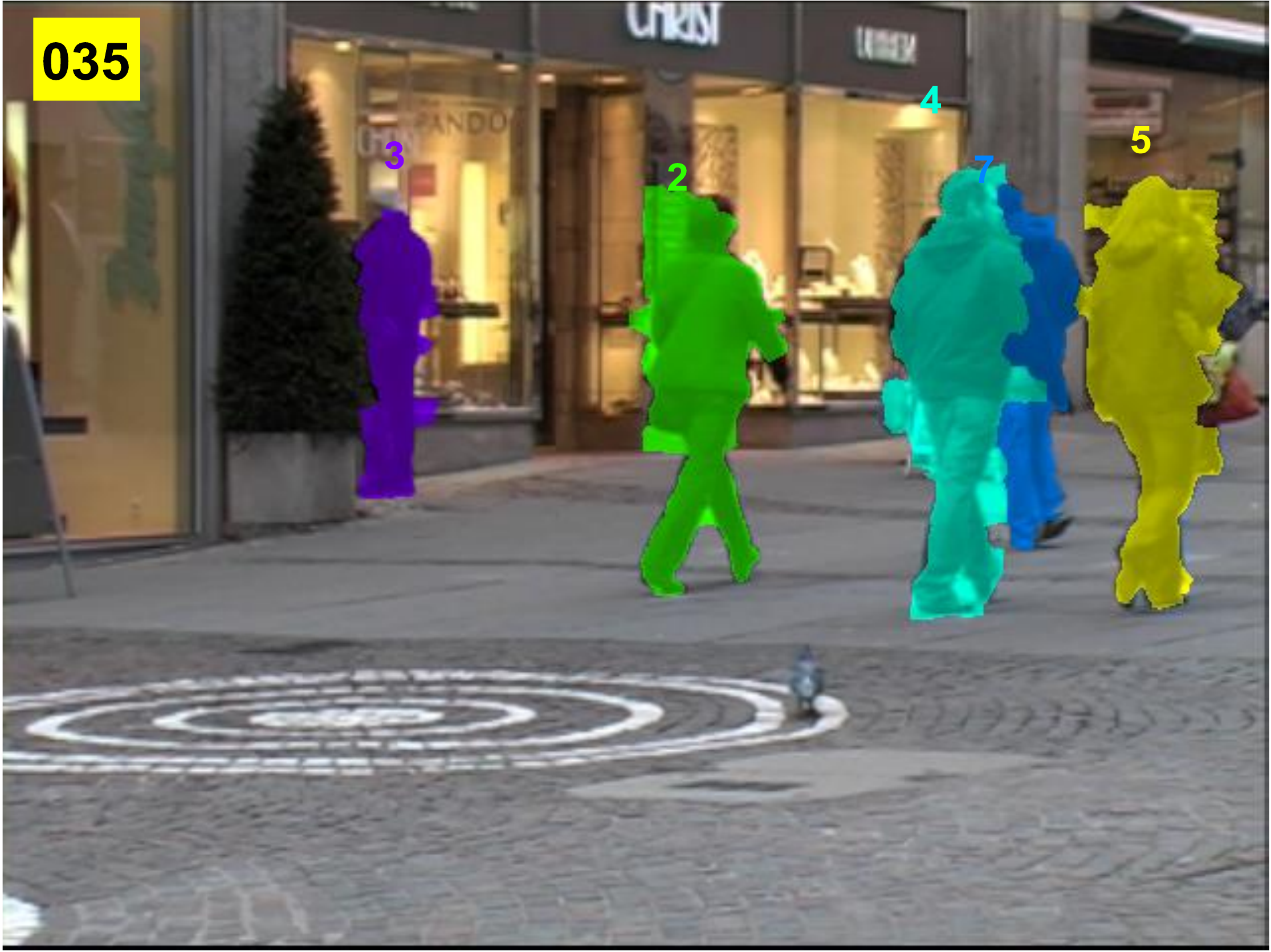}}%
	\subfigure{
		\includegraphics[width=0.24\textwidth]{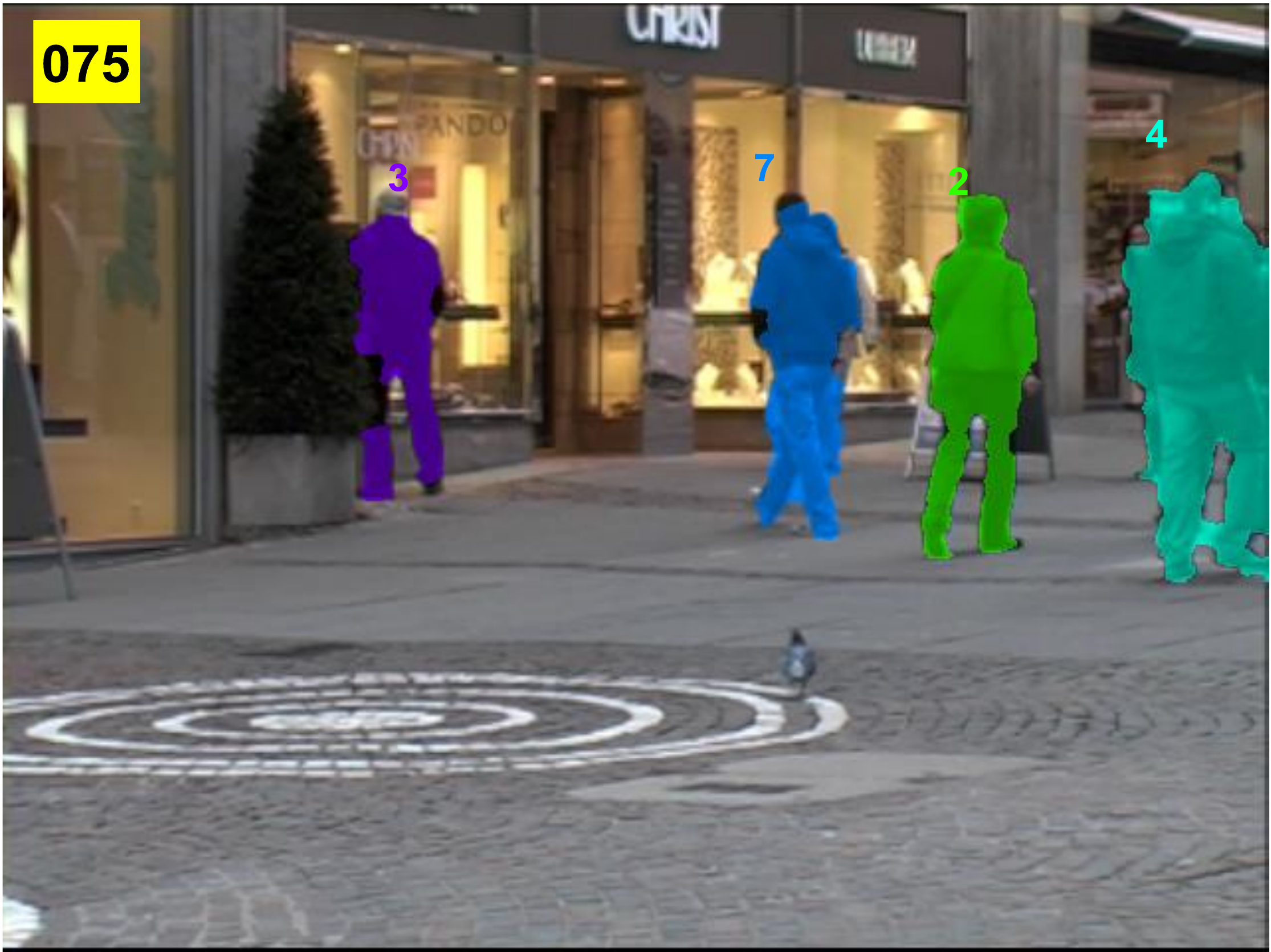}}%
	\subfigure{
		\includegraphics[width=0.24\textwidth]{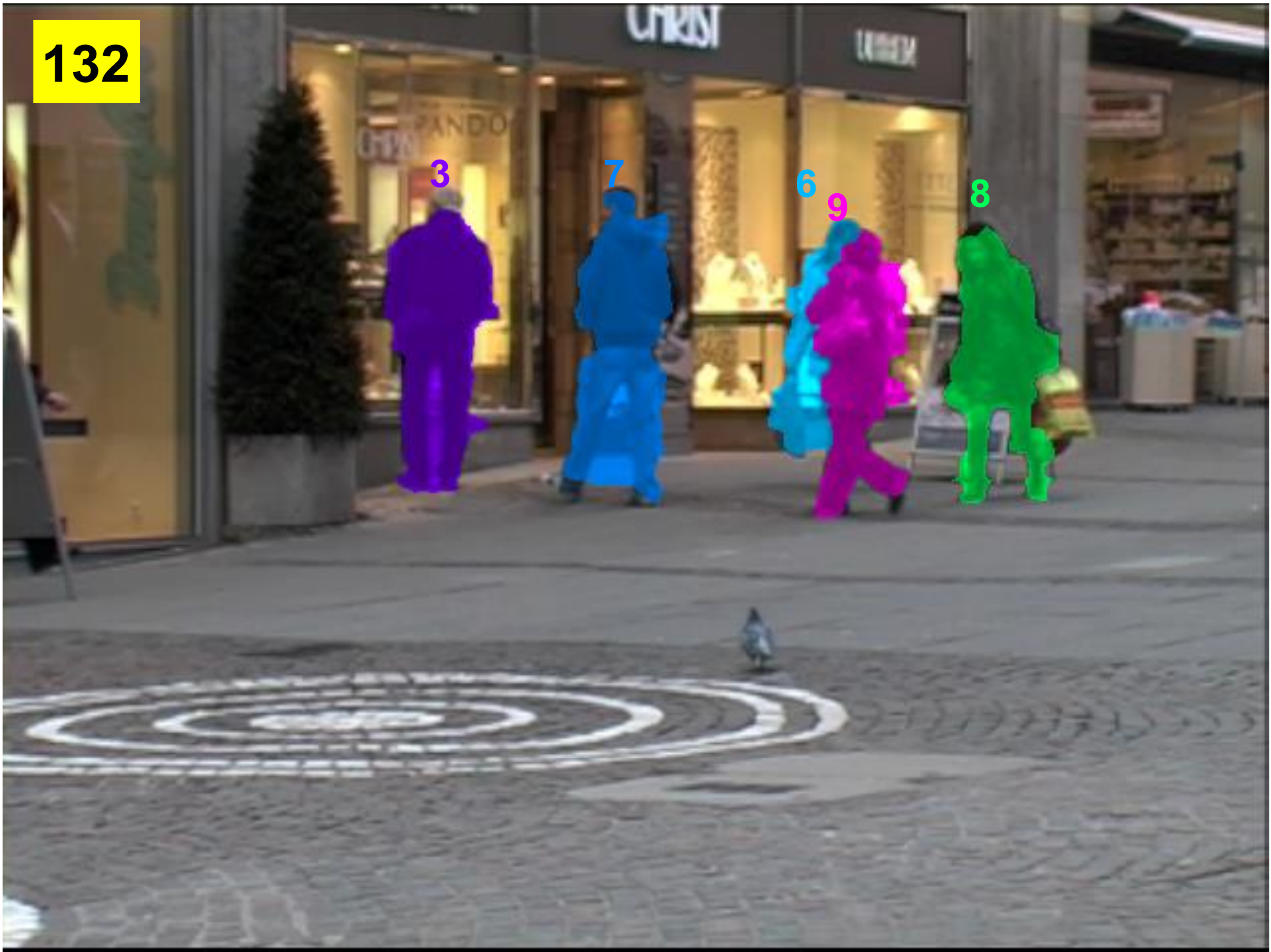}}%
	\subfigure{
		\includegraphics[width=0.24\textwidth]{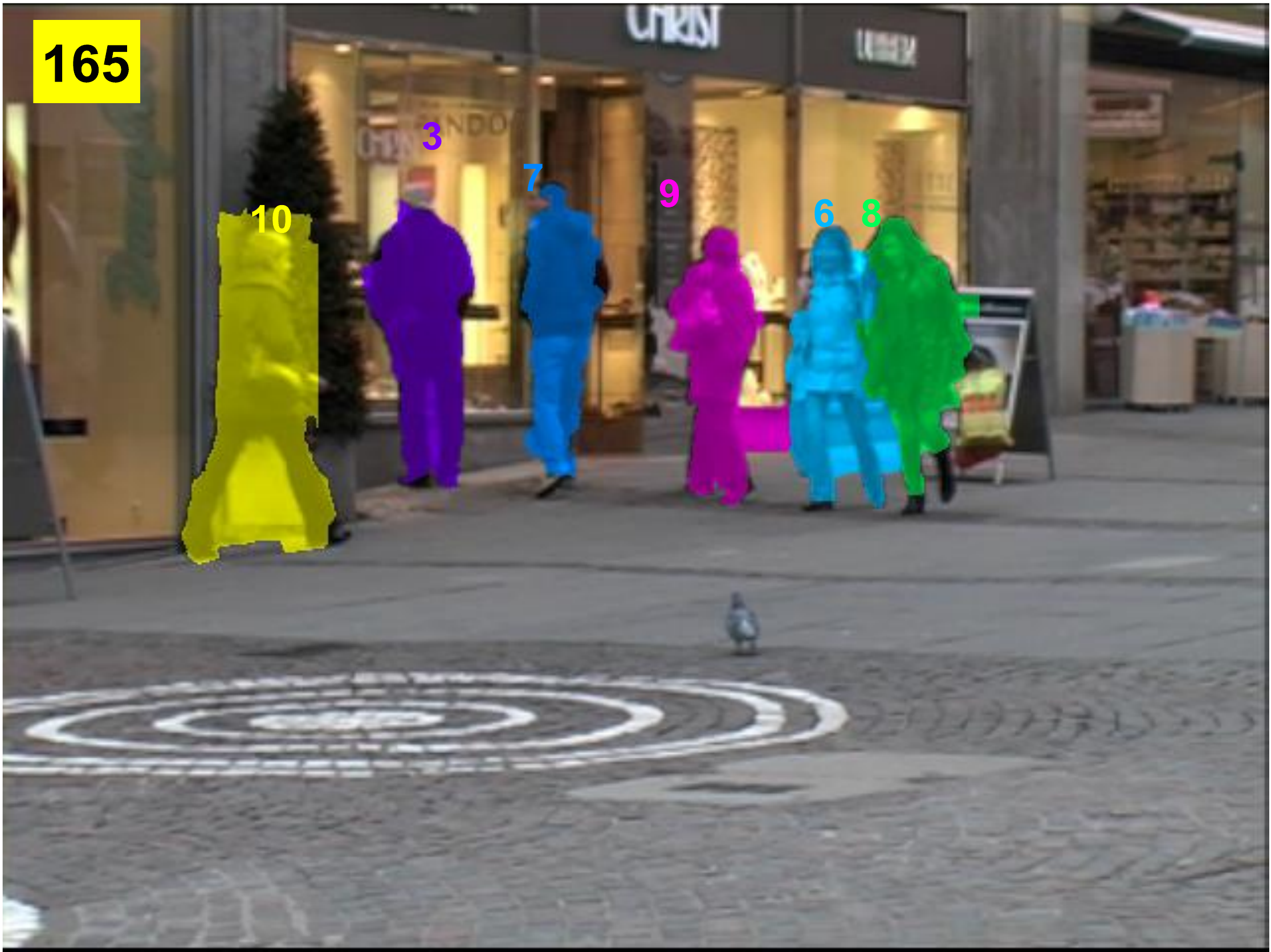}}	
	\caption{Examples of segmentation and tracking results on PETS-S2L1, TUD-Crossing and TUD-Stadtmitte. Each target is shown by a unique color.}
	\label{fig:fig4}
\end{figure}

Moreover, we show number of extracted objects with varying threshold $\alpha$ on ratio of correctly labelled pixels per ground truth mask in Fig. \ref{fig:fig6}. An object is extracted if more than $\alpha$ of its ground truth mask is correctly covered. Our approach is able to extract more objects for all different thresholds compared to previous methods and ``Ours - Seg".

\subsection{Tracking}
To quantitatively evaluate the tracking performance of our approach, both popular CLEAR MOT metrics \cite{bernardin2008evaluating} and Trajectory Based Metrics (TBM) \cite{wu2006tracking} are employed. MOTA considers the number of misses, false positives and mismatches, while MOTP measures the estimated object locations accuracy. Rcll, Prcn, MT, ML, Frag and IDS respectively are recall, precision, percentage of mostly tracked trajectories, percentage of mostly lost trajectories, number of trajectory fragments and number of identity switches.

Table \ref{tab:tab1} shows the quantitative comparison of tracking performance with previous methods and variants of our approach. Our approach outperforms state-of-the-art multiple target trackers. In particular, the number of ID-switches is substantially reduced compared to other methods. By incorporating segmentation, ``Ours" improves the results of ``Ours - Track" by a large margin, indicating that segmentation helps tracker avoid drifting and ID-switch. In addition, ``Ours - Auto" achieves similar results as ``Ours". As explained before, some targets may be initialized late using human detections, resulting in more false negatives and the small drop in MOTA.

\begin{table}
	\scriptsize
	\begin{center}
		\begin{tabu} to \textwidth{|X[2.25,c]|X[3,c]|X[0.9,c]|X[0.9,c]|X[0.6,c]|X[0.6,c]|X[0.6,c]|X[0.6,c]|X[0.6,c]|X[0.6,c]|X[0.6,c]}
			\hline
			Dataset & Method & MOTA & MOTP & Rcll & Prcn & MT & ML & Frag & IDS \\
			\hline
			\multirow{6}{*}{PETS-S2L1}
			& Milan et al. \cite{milan2015joint} & 85.3 & 77.5 & 98.1 & 88.7 & 100 & 0 & 24 & 9 \\
			& Chari et al. \cite{chari2015pairwise} & 85.5 & 76.2 & 92.4 & 94.3 & 94.7 & 0 & 74 & 56 \\
			& Zamir et al. \cite{zamir2012gmcp} & 90.3 & 69.0 & 93.6 & 96.5 & - & - & - & 8 \\
			& \textbf{Ours - Track} & 12.3 & 73.3 & 62.6 & 57.6 & 52.6 & 5.3 & 20 & 8 \\
			& \textbf{Ours - Auto} & 88.5 & 67.9 & 92.9 & 93.9 & 89.5 & 0 & 2 & 0 \\
			& \textbf{Ours} & 92.5 & 68.2 & 98.0 & 97.9 & 94.7 & 0 & 2 & 0 \\
			\hline
			\multirow{7}{*}{TUD-Crossing}
			& Milan et al. \cite{milan2015joint} & 59.2 & 73.1 & 83.9 & 77.4 & 66.7 & 0 & 21 & 8 \\
			& Breitenstein et al. \cite{breitenstein2011online} & 84.3 & 71.0 & -& - & - & - & - & 2 \\
			& Brendel et al. \cite{brendel2011multiobject} & 85.9 & 73.0 & - & - & - & - & - & 2 \\
			& Zamir et al. \cite{zamir2012gmcp} & 91.6 & 75.6 & 98.6 & 92.8 & - & - & - & 0 \\
			& \textbf{Ours - Track} & 79.2 & 62.3 & 86.1 & 91.8 & 83.3 & 0 & 1 & 0 \\
			& \textbf{Ours - Auto} & 88.1 & 73.5 & 91.0 & 97.0 & 91.7 & 0 & 1 & 0 \\
			& \textbf{Ours} & 91.7 & 62.4 & 94.7 & 96.5 & 91.7 & 0 & 1 & 0 \\
			\hline
			\multirow{6}{*}{TUD-Stadtmitte}
			& Milan et al. \cite{milan2015joint} & 68.0 & 55.9 & 74.4 & 94.3 & 60.0 & 0 & 3 & 3 \\
			& Chari et al. \cite{chari2015pairwise} & 51.6 & 61.6 & 59.6 & 89.9 & 20.0 & 0 & 22 & 15 \\
			& Zamir et al. \cite{zamir2012gmcp} & 77.7 & 63.4 & 95.6 & 81.4 & - & - & - & 0 \\
			& \textbf{Ours - Track} & 51.3 & 78.9 & 68.1 & 80.3 & 70.0 & 0 & 2 & 2 \\
			& \textbf{Ours - Auto} & 81.7 & 76.5 & 82.6 & 98.2 & 70.0 & 0 & 0 & 0 \\
			& \textbf{Ours} & 83.8 & 78.7 & 84.5 & 98.0 & 80.0 & 0 & 0 & 0 \\
			\hline
		\end{tabu}
	\end{center}
	\caption{\label{tab:tab1}A comparison of tracking results on PETS-S2L1, TUD-Crossing and TUD-Stadtmitte.}
\end{table}

\section{Conclusion}
We present a novel framework that combines online multiple target tracking and segmentation in a video. The two tasks are closely related and solving them should help each other. Tracking and segmentation are jointly optimized using dual decomposition, which leads to more accurate segmentation results and also helps resolve typical difficulties in tracking, such as occlusion handling, ID-switch and track drifting. Moreover, more detailed representation of targets - pixel-level target foreground labeling, is obtained rather than coarse bounding boxes.

\clearpage

\bibliographystyle{splncs}
\bibliography{egbib.bib}

\begin{thebibliography}{10}

\bibitem{zamir2012gmcp}
Zamir, A.R., Dehghan, A., Shah, M.:
\newblock Gmcp-tracker: Global multi-object tracking using generalized minimum
  clique graphs.
\newblock In: ECCV.
\newblock (2012)

\bibitem{zhang2008global}
Zhang, L., Li, Y., Nevatia, R.:
\newblock Global data association for multi-object tracking using network
  flows.
\newblock In: CVPR. (2008)

\bibitem{pirsiavash2011globally}
Pirsiavash, H., Ramanan, D., Fowlkes, C.C.:
\newblock Globally-optimal greedy algorithms for tracking a variable number of
  objects.
\newblock In: CVPR. (2011)

\bibitem{zhang2013structure}
Zhang, L., Maaten, L.:
\newblock Structure preserving object tracking.
\newblock In: CVPR. (2013)

\bibitem{dehghan2015target}
Dehghan, A., Tian, Y., Torr, P.H., Shah, M.:
\newblock Target identity-aware network flow for online multiple target
  tracking.
\newblock In: CVPR. (2015)

\bibitem{berclaz2011multiple}
Berclaz, J., Fleuret, F., T{\"u}retken, E., Fua, P.:
\newblock Multiple object tracking using k-shortest paths optimization.
\newblock PAMI (2011)

\bibitem{butt2013multi}
Butt, A.A., Collins, R.T.:
\newblock Multi-target tracking by lagrangian relaxation to min-cost network
  flow.
\newblock In: CVPR. (2013)

\bibitem{brox2010object}
Brox, T., Malik, J.:
\newblock Object segmentation by long term analysis of point trajectories.
\newblock In: ECCV.
\newblock (2010)

\bibitem{ma2012maximum}
Ma, T., Latecki, L.J.:
\newblock Maximum weight cliques with mutex constraints for video object
  segmentation.
\newblock In: CVPR. (2012)

\bibitem{zhang2013video}
Zhang, D., Javed, O., Shah, M.:
\newblock Video object segmentation through spatially accurate and temporally
  dense extraction of primary object regions.
\newblock In: CVPR. (2013)

\bibitem{yin2009shape}
Yin, Z., Collins, R.T.:
\newblock Shape constrained figure-ground segmentation and tracking.
\newblock In: CVPR. (2009)

\bibitem{li2013video}
Li, F., Kim, T., Humayun, A., Tsai, D., Rehg, J.:
\newblock Video segmentation by tracking many figure-ground segments.
\newblock In: ICCV. (2013)

\bibitem{wen2015jots}
Wen, L., Du, D., Lei, Z., Li, S.Z., Yang, M.H.:
\newblock Jots: Joint online tracking and segmentation.
\newblock In: CVPR. (2015)

\bibitem{bibby2010real}
Bibby, C., Reid, I.:
\newblock Real-time tracking of multiple occluding objects using level sets.
\newblock In: CVPR. (2010)

\bibitem{horbert2011level}
Horbert, E., Rematas, K., Leibe, B.:
\newblock Level-set person segmentation and tracking with multi-region
  appearance models and top-down shape information.
\newblock In: ICCV. (2011)

\bibitem{mitzel2010multi}
Mitzel, D., Horbert, E., Ess, A., Leibe, B.:
\newblock Multi-person tracking with sparse detection and continuous
  segmentation.
\newblock In: ECCV.
\newblock (2010)

\bibitem{milan2015joint}
Milan, A., Leal-Taix{\'e}, L., Schindler, K., Reid, I.:
\newblock Joint tracking and segmentation of multiple targets.
\newblock In: CVPR. (2015)

\bibitem{henriques2012exploiting}
Henriques, J.F., Caseiro, R., Martins, P., Batista, J.:
\newblock Exploiting the circulant structure of tracking-by-detection with
  kernels.
\newblock In: ECCV.
\newblock (2012)

\bibitem{strandmark2010parallel}
Strandmark, P., Kahl, F.:
\newblock Parallel and distributed graph cuts by dual decomposition.
\newblock In: CVPR. (2010)

\bibitem{wu2012coupling}
Wu, Z., Thangali, A., Sclaroff, S., Betke, M.:
\newblock Coupling detection and data association for multiple object tracking.
\newblock In: CVPR. (2012)

\bibitem{wang2011multi}
Wang, H., Koller, D.:
\newblock Multi-level inference by relaxed dual decomposition for human pose
  segmentation.
\newblock In: CVPR. (2011)

\bibitem{tsochantaridis2005large}
Tsochantaridis, I., Joachims, T., Hofmann, T., Altun, Y.:
\newblock Large margin methods for structured and interdependent output
  variables.
\newblock In: The Journal of Machine Learning Research. (2005)

\bibitem{dalal2005histograms}
Dalal, N., Triggs, B.:
\newblock Histograms of oriented gradients for human detection.
\newblock In: CVPR. (2005)

\bibitem{domke2006deformation}
Domke, J., Aloimonos, Y.:
\newblock Deformation and viewpoint invariant color histograms.
\newblock In: BMVC. (2006)

\bibitem{crammer2006online}
Crammer, K., Dekel, O., Keshet, J., Shalev-Shwartz, S., Singer, Y.:
\newblock Online passive-aggressive algorithms.
\newblock The Journal of Machine Learning Research (2006)

\bibitem{rother2004grabcut}
Rother, C., Kolmogorov, V., Blake, A.:
\newblock Grabcut: Interactive foreground extraction using iterated graph cuts.
\newblock In: ACM transactions on graphics (TOG). (2004)

\bibitem{achanta2012slic}
Achanta, R., Shaji, A., Smith, K., Lucchi, A., Fua, P., Susstrunk, S.:
\newblock Slic superpixels compared to state-of-the-art superpixel methods.
\newblock PAMI (2012)

\bibitem{boykov2001fast}
Boykov, Y., Veksler, O., Zabih, R.:
\newblock Fast approximate energy minimization via graph cuts.
\newblock PAMI (2001)

\bibitem{ferryman2010pets2010}
Ferryman, J., Ellis, A.:
\newblock Pets2010: Dataset and challenge.
\newblock In: AVSS. (2010)

\bibitem{andriluka2008people}
Andriluka, M., Roth, S., Schiele, B.:
\newblock People-tracking-by-detection and people-detection-by-tracking.
\newblock In: CVPR. (2008)

\bibitem{andriluka2010monocular}
Andriluka, M., Roth, S., Schiele, B.:
\newblock Monocular 3d pose estimation and tracking by detection.
\newblock In: CVPR. (2010)

\bibitem{breitenstein2011online}
Breitenstein, M.D., Reichlin, F., Leibe, B., Koller-Meier, E., Van~Gool, L.:
\newblock Online multiperson tracking-by-detection from a single, uncalibrated
  camera.
\newblock PAMI (2011)

\bibitem{felzenszwalb2010object}
Felzenszwalb, P.F., Girshick, R.B., McAllester, D., Ramanan, D.:
\newblock Object detection with discriminatively trained part-based models.
\newblock PAMI (2010)

\bibitem{bernardin2008evaluating}
Bernardin, K., Stiefelhagen, R.:
\newblock Evaluating multiple object tracking performance: the clear mot
  metrics.
\newblock Journal on Image and Video Processing (2008)

\bibitem{wu2006tracking}
Wu, B., Nevatia, R.:
\newblock Tracking of multiple, partially occluded humans based on static body
  part detection.
\newblock In: CVPR. (2006)

\bibitem{chari2015pairwise}
Chari, V., Lacoste-Julien, S., Laptev, I., Sivic, J.:
\newblock On pairwise costs for network flow multi-object tracking.
\newblock In: CVPR. (2015)

\bibitem{brendel2011multiobject}
Brendel, W., Amer, M., Todorovic, S.:
\newblock Multiobject tracking as maximum weight independent set.
\newblock In: CVPR. (2011)

\end{thebibliography}
\end{document}